\documentclass[%
 reprint,
 superscriptaddress,
 amsmath,amssymb,
 aps,
 nofootinbib,
]{revtex4-2}

\usepackage{graphicx,xcolor}
\usepackage{dcolumn}
\usepackage{bm}
\usepackage{listings}
\usepackage{orcidlink}
\usepackage[normalem]{ulem}
\usepackage{tikz}
\usetikzlibrary{positioning, arrows.meta, calc}

\begin{document}

\title{Generalist Vision-Language Models for Fast Radio Burst detection: a zero-shot benchmark against a specialized detector}

\author{Raiff H. Santos\,\orcidlink{0009-0009-4616-3183}}
\email{raiffhugos@gmail.com}
\affiliation{Unidade Acadêmica de Física, \\
Universidade Federal de Campina Grande,\\ 
R. Aprígio Veloso 882, Bairro Universitário, 58429-900, Campina Grande, PB, Brazil}

\author{Amilcar R. Queiroz\,\orcidlink{0000-0002-4785-5589}}
\email{amilcarq@df.ufcg.edu.br}
\affiliation{Unidade Acadêmica de Física, \\
Universidade Federal de Campina Grande,\\ 
R. Aprígio Veloso 882, Bairro Universitário, 58429-900, Campina Grande, PB, Brazil}

\author{Tharcisyo S. S. Duarte\,\orcidlink{0000-0002-6164-1915}}
\email{tharcisyo.duarte@ufca.edu.br}
\affiliation{Instituto de Formação de Educadores, Universidade Federal do Cariri,\\
Rua Olegário Emídio de Araújo, s/n, Aldeota, 63260-000, Brejo Santo, CE, Brazil}
\affiliation{Universidade Federal de Campina Grande,\\ 
R. Aprígio Veloso 882, Bairro Universitário, 58429-900, Campina Grande, PB, Brazil}

\author{K. E. L. de Farias\,\orcidlink{0000-0002-9418-9566}}
\email{klecio.lima@uaf.ufcg.edu.br}
\affiliation{Unidade Acadêmica de Física, \\
Universidade Federal de Campina Grande,\\ 
R. Aprígio Veloso 882, Bairro Universitário, 58429-900, Campina Grande, PB, Brazil}
\affiliation{Centre of Excellence ENSEMBLE3 Sp. z o. o., Wolczyńska Str. 133, 01-919, Warsaw, Poland}

\author{Rafael A. Batista\,\orcidlink{0000-0002-7906-1505}}
\email{rafael.batista@uaf.ufcg.edu.br}
\affiliation{Unidade Acadêmica de Física, \\
Universidade Federal de Campina Grande,\\ 
R. Aprígio Veloso 882, Bairro Universitário, 58429-900, Campina Grande, PB, Brazil}

\date{\today}

\begin{abstract}
Fast Radio Burst (FRB) detection increasingly relies on specialized deep learning models that require large task-specific training sets and cannot be redefined without retraining. We evaluate whether small, open-weight, locally run generalist Vision-Language Models (VLMs) can detect FRBs in dynamic spectra under a zero-shot, prompt-only regime. On a balanced binary benchmark of 2000 simulated L-band spectra, Gemma 4 E2B reaches an accuracy of 94.05\%, statistically indistinguishable from the specialized detector SwinYNet (92.85\%), with a far lower false-positive rate on structured RFI (4.8\% vs. 24.6\%) and none on pure noise, though SwinYNet ranks perfectly (ROC-AUC 1.0000 vs. 0.9520). Rewriting the prompt alone reconfigures the same models for three-class FRB/RFI/noise classification, reaching up to 86.0\% accuracy without a single false FRB while classifying each 2 s spectrum in 1.0--1.5 s, faster than the observation itself. Applied unchanged to the 1600 real FAST observations of FAST-FREX, they reject real interference almost perfectly (2 and 5 false positives in 1000 negatives) but recover only 28.5\% and 27.0\% of the 600 catalogued bursts, against 95.7\% reported for SwinYNet on the same files. Stratifying those bursts by the dispersed signal in the image shows the limit to be the input representation rather than the classifier, recall rising to 84--85\% where the sweep is unambiguous and collapsing to 1\% on the 13\% of positives carrying no detectable signal in a 2 s undedispersed full-band view. The simulated bursts are nearly 30 times brighter in median, and at matched brightness the recalls agree to within a few points.

\end{abstract}

\maketitle


\section{Introduction}

Fast Radio Bursts (FRBs) are intense radio transients with durations of only a few milliseconds and predominantly of extragalactic origin. Since the discovery of the first FRB by~\cite{lorimer2007bright}, these events have become one of the most intriguing topics in modern astrophysics and cosmology \cite{Katz2016,Petroff2019, Zhang2020a, Zhang2023, Rajwade2024, Ma2025}. FRBs are characterized by their large dispersion measures (DMs) \cite{Petroff:2014taa,Petroff2019,Cordes:2019cmq}, which generally exceed the expected contribution from the Milky Way, indicating propagation through the ionized intergalactic medium (IGM). Over the past years, hundreds of FRBs have been detected by several radio observatories, including the Parkes telescope, CHIME \cite{bandura2014canadian}, ASKAP \cite{johnston2007science}, FAST \cite{nan2011five}, and MeerKAT \cite{camilo2018revival}, revealing a diverse population of repeating and apparently non-repeating sources. That division has itself become a machine-learning problem, with unsupervised analyses of the CHIME/FRB catalogs separating the two classes mainly by spectral morphology, the spectral running and spectral index emerging as the dominant features~\cite{sun2025exploring,sun2026practical,sun2026unveiling}. Because of their extragalactic characteristics, they have recently been used for cosmological probes in several scenarios \cite{Lemos:2022kdh,Fortunato:2023deh,Lemos:2025qyh,Sales:2025shu,Sales:2026smc,Santos:2026qnl,Ribeiro:2026lyd}.

Processing these streams in real-time requires automated detection pipelines. Traditional pipelines rely on matched filtering and dedispersion, which are computationally expensive \cite{Rajwade2024, Cao2025, Yang2025anomalies,Yang2023classifying}. Historically, this process was dominated by CPU-based software packages like PRESTO (PulsaR Exploration and Search TOolkit) \cite{ransom2001, ransom2011presto}, which established the foundational algorithms for incoherent dedispersion and candidate folding \cite{lorimer2005handbook}. To cope with the escalating data rates of modern radio telescopes, the field transitioned toward GPU-accelerated architectures, such as the Heimdall pipeline \cite{barsdell2012}, which enabled high-throughput single-pulse searches in real-time. More recently, the paradigm has shifted toward mitigating the computational burden of brute-force dedispersion altogether by introducing visual and morphological feature extraction \cite{BelmonteDiaz2026}. A key example is RaSPDAM (Radio Single-Pulse Detection Algorithm Based on Visual Morphological Features) \cite{guo2025accelerating}, an AI-driven framework that treats the dynamic spectrum as a computer vision problem to rapidly identify transient candidates. Extending this visual approach to its current state-of-the-art, advanced Vision Transformer (ViT) architectures \cite{dosovitskiy2021image} have established a powerful new paradigm for end-to-end processing. Specifically, models like SwinYNet \cite{chen2026swinynet} represent a major shift in FRB search by performing multi-task learning that simultaneously combines real-time burst detection, pixel-level signal segmentation, and direct parameter estimation from raw dynamic spectra without any prior dedispersion. While these specialized deep learning models achieve exceptional performance, they traditionally rely on large, specialized astronomical data sets for training. As an alternative to training such dedicated models from scratch, the rapid advancement of Multimodal Large Language Models (MLLMs) \cite{yin2024mllm_survey}, or Vision-Language Models (VLMs), has introduced a parallel paradigm where generalist foundation models can be applied directly to visual classification tasks using only natural language prompts.

This paradigm has already begun to be explored in astronomical problems, including galaxy morphology classification~\cite{he2026astrosight, tanoglidis2026vision} and domain-adapted astronomical assistants~\cite{zaman2025astrollava, kamai2026talking}. In optical transients, it was shown in Ref.~\cite{stoppa2025textual} that a generalist Large Language Model (LLM), Google's Gemini~\cite{team2023gemini}, can approach the performance of a convolutional neural network in classifying real sources versus artifacts (\emph{real versus bogus}) across three surveys (Pan-STARRS, MeerLICHT, and ATLAS) with the advantage of generating human-readable descriptions for each candidate from a few examples and concise instructions. That same work, however, points to an important practical limitation. Employing a large LLM through a commercial API is, at present, infeasible at the scale of upcoming surveys (such as the Vera Rubin Observatory, which is expected to generate on the order of 10 million alerts per night), both because of token-generation latency and prohibitive costs. The authors indicate as promising directions the fine-tuning of smaller open-source LLMs, with one to two billion parameters, and model quantization to enable faster local inference. It is precisely in this direction that models such as Gemma 4 have advanced, providing open, small, locally executable VLMs~\cite{gemmateam2026gemma4}, the central premise of the approach we adopt here.

Applications to radio data have also begun to appear, although still relying on fine-tuning and outside the domain of FRBs. In Ref.~\cite{zhao2024pulsar}, STARWHISPER-PULSAR was proposed, which fine-tunes an MLLM over visual, textual, and numerical modalities to distinguish pulsar candidates from noise, outperforming specialized machine-learning methods. In Ref.~\cite{zhao2025unveiling}, the RadioAstroVQA set was built and MLLMs were fine-tuned for image classification and visual question answering (VQA) in radio astronomy, achieving accuracy comparable to or higher than that of existing deep-learning models. Similarly, in Ref.~\cite{riggi2025radio}, a VLM, radio-llava, was fine-tuned for the analysis and classification of radio-astronomical sources. On the other hand, in Ref.~\cite{drozdova2025radio}, the authors evaluated generalist VLMs (Qwen and Gemini) on the morphological classification of radio galaxies (MiraBest FR-I/FR-II). Prompt-only strategies already achieve good performance, but the outputs are unstable, varying strongly with superficial changes to the prompt, while a light fine-tuning (LoRA) rivals specialized models. Taken together, these works confirm the potential of MLLMs and VLMs in astronomy, but they typically resort to fine-tuning, to large models accessed through an API, or to tasks distinct from FRB detection.

In this paper, in contrast, we present a systematic evaluation of small, local VLMs (Gemma 4 E2B and E4B) applied to the task of FRB detection. We construct a controlled, simulated benchmark of 3000 L-band dynamic spectra, from which we draw a balanced binary subset of 2000 samples, using \texttt{simulateSearch}~\cite{Luo2022_simulateSearch_paper} containing FRB signals, various classes of structured Radio Frequency Interference (RFI), and background noise. We compare the zero-shot, prompt-only performance of these VLMs against SwinYNet, a state-of-the-art specialized deep learning detector. We analyze their performance across binary and multiclass tasks, evaluate their behavior as probabilistic classifiers, characterize their failure modes on faint signals, and discuss the computational trade-offs of deploying generalist models in operational astronomical pipelines. We then carry the same models, unchanged, to real observations, evaluating them on the full FAST-FREX data set~\cite{guo2025accelerating} of 1600 FAST files, separating what the classifier fails to recognize from what its input representation does not render visible in the first place, and placing the result alongside the published performance of specialized detectors on the same data set.

This paper is structured as follows. Section~\ref{sec:frb} reviews the physics of Fast Radio Bursts and their dispersive signature, the RFI and noise that contaminate the search, and the traditional single-pulse pipelines used to detect them. Section~\ref{sec:vlm-background} introduces multimodal models and the class of VLMs evaluated in this work. Section~\ref{sec:simulation} details the simulation setup and benchmark data set. Section~\ref{sec:vlm} describes the zero-shot VLM classification protocol and prompt design. Section~\ref{sec:results} presents the quantitative results, paired comparison with SwinYNet, and error analyses. Section~\ref{sec:real} validates the same models on real FAST observations. Finally, Section~\ref{sec:conclusions} summarizes our findings and outlines directions for future work.

\section{Fast Radio Bursts, RFI, and traditional searches}\label{sec:frb}

This section provides the physical and operational background for the classification task studied in this work. The goal is not a complete review of FRB astrophysics or of transient-search software, but a concise account of three points on which the rest of the paper depends: what an FRB looks like in a dynamic spectrum, what the two contaminants a search must reject, viz. RFI and noise, look like in the same representation, and how a traditional pipeline turns a raw data stream into candidates. The detection task is fundamentally binary: an FRB must be separated from everything that is not an FRB, whether structured interference or background noise. This is the decision that both the specialized detector and the generalist VLM perform in our comparison. The VLM, however, is not restricted to this binary decision. It can also be prompted to go beyond it and separate RFI from noise as well, a possibility we return to later in Sect.~\ref{sec:multiclass-task}.

\subsection{The dispersed burst and its visual signature}

A Fast Radio Burst is a radio transient of millisecond duration, high energy, and predominantly extragalactic origin~\cite{lorimer2007bright}, although a bright radio burst from the Galactic magnetar SGR~1935+2154 has tied at least some FRBs to magnetars~\cite{Bocheneketal2020}. As the broadband pulse propagates through the ionized plasma of the interstellar and intergalactic media, its lower frequencies are progressively delayed relative to the higher ones, an effect known as dispersion~\cite{lorimer2005handbook}. The resulting delay between two frequencies $\nu_{\mathrm{lo}}$ and $\nu_{\mathrm{hi}}$ is set by the DM through\footnote{The prefactor is written in Gaussian (CGS) units. For DM in pc cm$^{-3}$ and frequencies in MHz it evaluates to the dispersion constant $D \approx 4.15 \times 10^{3}$ MHz$^{2}$ pc$^{-1}$ cm$^{3}$ s, the value used throughout the analysis of Sect.~\ref{sec:real}.}
\begin{equation}
\Delta t=\frac{e^2}{2 \pi m_e c}\left(\nu_{\mathrm{lo}}^{-2}-\nu_{\mathrm{hi}}^{-2}\right) \mathrm{DM}\,,
\end{equation}
and 
\begin{equation}
\mathrm{DM} = \int_0^{d} n_e(l)\, dl,
\end{equation}
where $n_e(l)$ is the free-electron density along the path. The observed DM combines contributions from the Milky Way, the intergalactic medium, and the host galaxy~\cite{macquart2020census}: 
\begin{equation}
\mathrm{DM}_{\mathrm{obs}}(z)=\mathrm{DM}_{\mathrm{MW}}+\mathrm{DM}_{\mathrm{IGM}}+\frac{\mathrm{DM}_{\text {host }}}{1+z}\,.
\end{equation}
A DM in excess of the expected Galactic contribution~\cite{yao2017new, cordes2002ne2001} is one of the primary indicators of an extragalactic source. Other propagation effects (scattering, scintillation, Faraday rotation, and plasma lensing, among others) further modify the pulse, but dispersion is the effect that dominates its appearance and underlies its detection.

Because the delay scales as $\nu^{-2}$, an FRB traces a characteristic curved track in the dynamic spectrum, the time--frequency plane in which the horizontal axis is time, the vertical axis is frequency, and the pixel intensity is signal power. Lower frequencies arrive systematically later, so the burst appears as a single, localized, plausibly broadband sweep whose curvature is set by the DM. This quadratic sweep is the visual signature a search must recover, and it is exactly the morphology the VLM is asked to identify from the image alone (Sect.~\ref{sec:vlm}). Dispersion is also what most cleanly separates an astrophysical burst from terrestrial interference, since the latter does not, in general, follow the same frequency-dependent delay~\cite{zhang2025drafts}. 

\subsection{Radio-frequency interference and noise}

The negative class of an FRB search is dominated by two very different phenomena: RFI and system noise. RFI is contamination of terrestrial or instrumental origin like, for instance, mobile networks, wireless links, radar, satellites, power lines, and the receiver electronics themselves. Because these sources sit far closer to the telescope than any astrophysical emitter, RFI is frequently orders of magnitude stronger than the signal of interest and can bury a genuine burst~\cite{zhang2025drafts}.

Crucially, RFI is not a single visual class. It appears as persistent narrowband bands (nearly horizontal lines at fixed frequencies), as broadband impulses with no dispersive delay (nearly vertical, zero-DM structures), as periodic or repetitive patterns, and as multiband or in-band occupancy from services such as wireless networks and satellites. What these morphologies share is the absence of the coherent $\nu^{-2}$ sweep of a dispersed burst. This makes them dangerous since some of them can partially imitate coherent structure and then survive as false candidates. System noise, by contrast, is the stochastic fluctuation of the instrument, with no coherent structure. It matters because dispersion spreads the burst energy over a wide time span and lowers its signal-to-noise ratio, letting faint bursts sink into the background. These two contaminants map directly onto the \texttt{RFI} and \texttt{NOISE} classes of our benchmark (Sect.~\ref{sec:simulation}), whose distinct behavior is what motivates reporting false positives separately for each (Sect.~\ref{sec:fp}).

\subsection{Traditional single-pulse searches}

Because the DM of a candidate is not known in advance, a traditional single-pulse search cannot look for the sweep directly; it must reconstruct the burst by testing many trial DMs~\cite{zhang2025drafts, guo2025accelerating}. After the strongest interference is flagged and masked in time and frequency, the search proceeds by dedispersion. For a given trial DM, each frequency channel is shifted back in time by the delay that this DM predicts, and the channels are summed into a single time series. When the trial DM matches the burst, the curved sweep re-aligns and the burst energy piles up into one narrow, high signal-to-noise (S/N) spike; when it does not, the channels add out of step and the energy smears back into the background. The two-dimensional search for a curved track is thereby reduced to a one-dimensional search for a peak, repeated over a dense grid of trial DMs. A genuine burst reaches its maximum S/N at its true DM and falls off on either side, whereas terrestrial interference, having no dispersive delay, peaks at zero DM; the trial DM that maximizes the response therefore both estimates the dispersion and helps distinguish the burst from RFI.

Each dedispersed time series must still be searched for a pulse of unknown duration. This is done by matched filtering. The series is correlated with boxcar templates of increasing width, and the S/N is largest when the template width matches the true burst width, since a narrower window discards signal while a wider one adds noise. Samples exceeding an S/N threshold, over the joint space of trial DM, arrival time, and boxcar width, are retained as single-pulse candidates. Because a real event produces detections at many neighboring trials, these are then clustered to collapse each event into a single candidate, which is finally labeled as astrophysical or as a false positive (historically by human inspection, and increasingly by machine-learning classifiers). This scheme underlies the standard search tools, from the CPU-based PRESTO~\cite{ransom2001, ransom2011presto} to the GPU-accelerated Heimdall~\cite{barsdell2012}, and it remains the dominant approach to FRB detection.

The cost of this procedure is dominated by dedispersion. Direct incoherent dedispersion scales as $O(N_t\,N_\nu\,N_{\mathrm{DM}})$ in the number of time samples $N_t$, frequency channels $N_\nu$, and trial DMs $N_{\mathrm{DM}}$; with hundreds to thousands of DM trials at high time and frequency resolution, a brute-force real-time search becomes very expensive~\cite{barsdell2012}. This burden has driven both the move to GPU acceleration and, more recently, a shift toward treating the dynamic spectrum as an image and applying computer-vision models to it directly, as in RaSPDAM~\cite{guo2025accelerating}, DRAFTS~\cite{zhang2025drafts}, and SwinYNet~\cite{chen2026swinynet}.

\subsection{Limitations of the traditional approach}

Beyond raw computational cost, the traditional pipeline has well-known operational limitations. The DM grid is redundant, so a single event is detected at many trial DMs and inflates the candidate list with duplicates. The output is dominated by false positives, most of them driven by RFI that survives the masking stage or is even introduced by it~\cite{zhang2025drafts}. Performance is highly sensitive to parameter choices, like the RFI-flagging algorithm, the spacing of the DM grid, the set of boxcar widths, and the S/N threshold, so that faint bursts near the detection limit are easily lost. And because the surviving candidates must still be separated from residual interference, the pipeline ultimately relies on human verification, which becomes the practical bottleneck as data volumes grow~\cite{chen2026swinynet}.

These limitations frame the rest of the paper. The RFI-driven false-positive rate and the human-inspection bottleneck are exactly the pressure points that learning-based methods try to relieve, and the fact that the FRB signature is, at bottom, a visual pattern in the dynamic spectrum is what makes an image-level approach natural. It is against this backdrop that we ask a narrower question: how much of that recognition is already within reach of a generalist VLM applied to the dynamic-spectrum image, with no domain-specific training at all.

\section{Multimodal models and VLMs}\label{sec:vlm-background}

This section characterizes the model class evaluated in this work and the operational regime adopted. The goal is not an exhaustive review of multimodal machine learning, but a concise account of how a generalist VLM, with no FRB-specific training, can be applied to dynamic-spectrum images, and how its performance should be interpreted as a lower bound on the potential of the approach.

\subsection{From language models to Vision-Language Models}

MLLMs extend LLMs to non-textual inputs~\cite{yin2024mllm_survey}. An LLM represents text as a sequence of \emph{tokens}, discrete units (typically subwords) that the model reads and generates. VLMs are the subclass that jointly conditions on an image $\mathbf{I}$ and a textual instruction $\mathbf{p}$ to produce a textual response $\mathbf{y}$. In the dominant connector architecture~\cite{liu2023llava}, a vision encoder $\mathcal{E}_{v}$ maps the image to visual tokens, a projector $W_{v}$ aligns them with the LLM embedding space, and an autoregressive decoder generates the answer:

\begin{equation}
\mathbf{z} = W_{v}\,\mathcal{E}_{v}(\mathbf{I}), \qquad
p(\mathbf{y}\mid \mathbf{I},\mathbf{p}) = \prod_{t=1}^{T} p\!\left(y_{t}\mid y_{<t}, \mathbf{z}, \mathbf{p}\right).
\label{eq:vlm-autoregressive}
\end{equation}

Here, $\mathbf{z}$ are the visual tokens already projected into the embedding space, and the response is generated token by token in an autoregressive way. Each token $y_{t}$ is produced conditioned on the previous tokens $y_{<t}$, the visual tokens $\mathbf{z}$, and the prompt $\mathbf{p}$, over the $T$ tokens of the response. Because the output is free text, the same structure can perform captioning, visual question answering, or, as in this work, systematic classification, depending solely on the prompt.

This generality, that is, a single model serving distinct tasks according to the instruction, comes from the pretraining regime. These models are trained on large collections of general-domain text and image-text pairs, which gives them broad visual and linguistic knowledge but does not specialize them in any particular scientific task. Throughout the paper, following the introduction, we use \emph{VLM} for this class of models applied to dynamic-spectrum images. The extent to which this general competence transfers to specialized scientific imagery is an open question, now actively probed by dedicated multimodal benchmarks in astronomy~\cite{ren2026systematic, chen2026astroalertbench, shi2025astrommbench}.

\subsection{Zero-shot classification and the prompt-only regime}\label{sec:zeroshot}

In this work, the VLMs operate in a zero-shot, prompt-only regime. \textit{Zero-shot} means that the model classifies each spectrum without having been exposed to any labeled example of the task, relying solely on the representations acquired during general multimodal pretraining and on a textual instruction that specifies the candidate classes. This use of natural language to reference visual concepts learned in pretraining, rather than a classifier head fitted to task-specific labels, is precisely the transfer mechanism introduced by CLIP~\cite{radford2021learning}, here applied to dynamic spectra. \textit{Prompt-only} means that no model weight is updated. The task is defined entirely in natural language.

This regime is therefore distinct from \textit{in-context} (few-shot) learning, in which a handful of labeled examples would be embedded in the prompt as demonstrations, a strategy adopted, for instance, in the classification of optical transients with LLMs~\cite{stoppa2025textual} and of radio galaxies with VLMs~\cite{drozdova2025radio}.

It is equally distinct from the other common strategies for adapting foundation models. There is no fine-tuning, i.e., no weight adjustment on FRB data; no retrieval-augmented generation (RAG), in which external documents would be injected into the prompt; and no agentic behavior or external tool calls. The model receives a single image and a single instruction, and returns a decision. This choice is deliberate, as it isolates the question of interest: how much of the ability to recognize an FRB is already latent in a generalist model, prior to any domain specialization? Accordingly, the performance reported here should be interpreted as a measure of potential, not as the ceiling the approach would reach with further adaptation.

\subsection{Generalist and specialized detectors}

The central difference between a generalist VLM and a detector such as
SwinYNet lies in the origin of its competence. SwinYNet is a supervised model built on the Swin Transformer backbone~\cite{liu2021swin}, trained end to end on FRB data to jointly perform detection, segmentation, and parameter estimation, and it operates on the PSRFITS files in its native domain. Formally, both approaches can be written as a composition of a visual feature extractor
$f_{\theta}$ and a task head $g_{\phi}$,

\begin{equation}
  \hat{y} = g_{\phi}\!\big(f_{\theta}(x)\big),
  \label{eq:detector_composition}
\end{equation}

but they differ in how the parameters $(\theta,\phi)$ are obtained. In the
specialized detector, both $\theta$ and $\phi$ are optimized on labeled FRB
data,

\begin{equation}
  (\theta^{\star},\phi^{\star})
  = \operatorname*{arg\,min}_{\theta,\phi}\,
  \mathcal{L}_{\mathrm{FRB}} .
  \label{eq:specialized_training}
\end{equation}

In the zero-shot VLM, $\theta$ is fixed from general multimodal pretraining and
the task head is induced by the textual prompt $P$ rather than fitted, so that
$\hat{y} = g_{P}\!\big(f_{\theta}(x)\big)$ with $\theta$ unchanged.

Applying the VLM to FRB detection is therefore a test of visual transfer:
checking whether a feature map $f_{\theta}$ learned in a general domain, encoding edges, elongated tracks, textures, and localized structures, is already expressive enough to separate a dispersed burst from interference and
from noise, without any update to $\theta$. This is the same hierarchical,
multiscale feature reuse that makes the Swin backbone an effective
general-purpose vision model~\cite{liu2021swin}, here probed in the opposite
regime, reused as-is, instead of fine-tuned.

This asymmetry is part of the study design and is examined in detail in the
methodological comparison (Sect.~\ref{sec:comparison}). It favors the
specialized detector in terms of access to information, which is precisely what
makes strong VLM performance interesting. The argument of the paper is not that
the VLM replaces a dedicated detector, but that a generalist tool already
recognizes the relevant patterns starting from almost no domain information,
supporting the potential thesis.

\subsection{Small, local, and open models}

A deliberate aspect of the study is the use of small, open-weight models that run locally, rather than large models accessed through a commercial API or the large, domain-specific foundation models now being trained for astronomy~\cite{parker2026aion}. This choice is enabled by a recent generation of compact open models, such as the Gemma family~\cite{gemmateam2024}, which are explicitly designed to be deployable across hardware ranging from consumer devices to servers, bringing state-of-the-art capability within reach of modest infrastructure.

The motivations are practical. Local execution removes the dependence on commercial services and per-call costs; it favors reproducibility, since a specific model version can be pinned and rerun; it keeps the data under the control of whoever performs the analysis; and it makes the approach accessible to groups without large computational infrastructure. Small models also bring the experiment closer to a realistic deployment scenario in radio astronomy, where candidate screening may need to happen near the instrument and under hardware constraints.

The cost of this choice is a per-model capability lower than that of the largest available systems. We argue that this works in favor of the analysis rather than against it. The performance reported here is therefore a \emph{conservative floor} for what the class of generalist VLMs could reach, since a larger model operating in the same prompt-only regime would be expected to match or exceed it. The two concrete models evaluated, their versions, and the hardware used are described in Sect.~\ref{sec:models}.

\subsection{Structured output and interpretability}

Because they produce text, VLMs can be instructed to respond in a structured format and, at the same time, to justify the decision in natural language. In this work, the model returns a label, a continuous probability that the image contains an FRB, a confidence, and a short justification accompanied by visual attributes, with the exact schema shown in Listing~\ref{lst:json}. This property is attractive for scientific use, because each decision comes with an inspectable explanation, something that specialized detectors rarely offer.

Even so, this justification should be treated as auditable auxiliary data, not as automatic physical proof. A model can produce a plausible explanation for an incorrect decision. Therefore, interpretability here is understood as inspectability of the decision, not as a guarantee of physical causality, a point taken up again among the threats to validity (Sect.~\ref{sec:threats}).

\section{Simulation and data set}\label{sec:simulation}

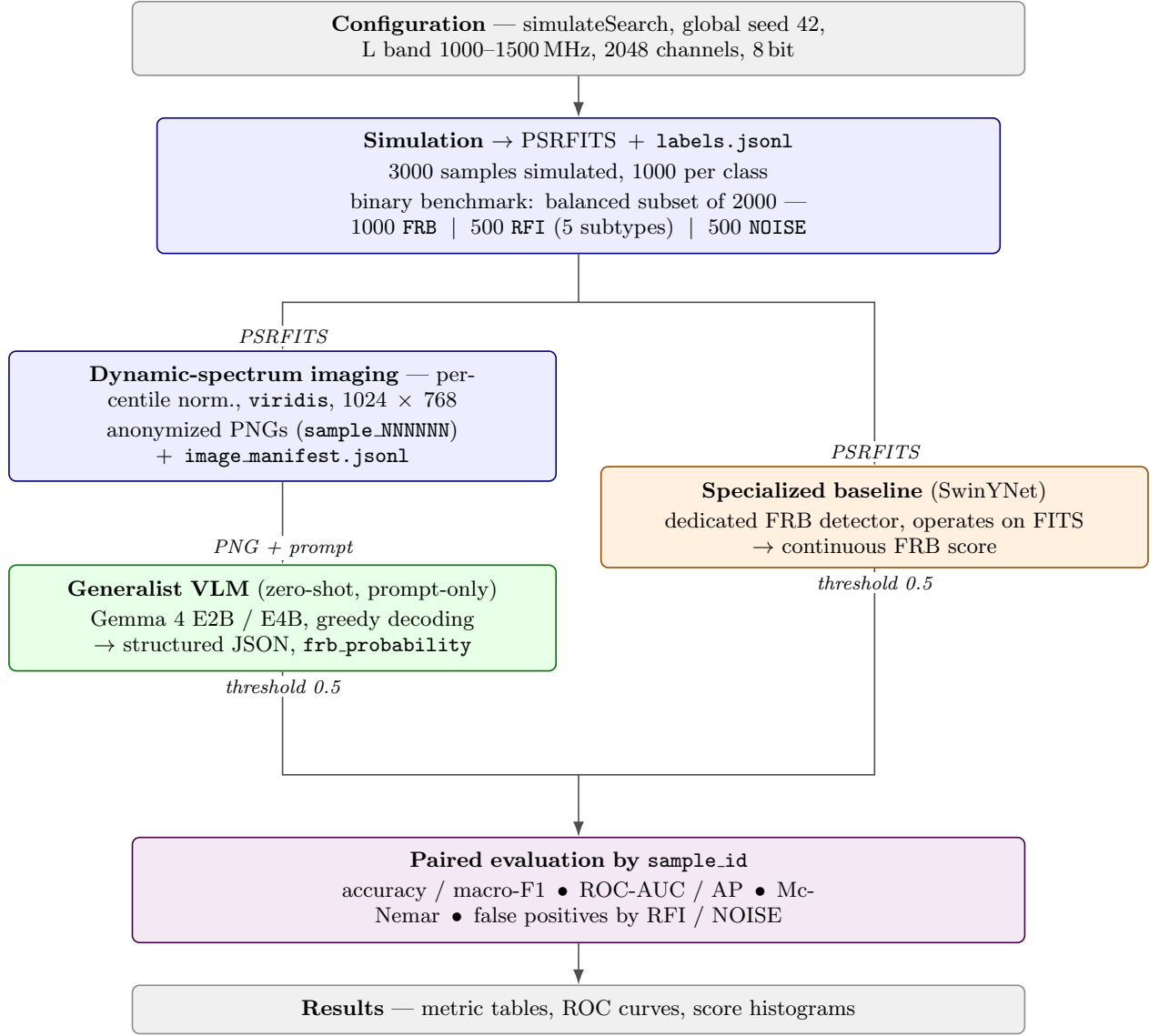
\begin{figure*}
\centering
\begin{tikzpicture}[
  font=\small,
  node distance=6mm,
  flow/.style={-{Latex[length=2.2mm,width=1.6mm]}, line width=0.6pt, draw=black!70},
  link/.style={line width=0.6pt, draw=black!70},
  base/.style={rounded corners=3pt, line width=0.6pt, align=center, inner sep=6pt},
  io/.style  ={base, draw=black!45,       fill=black!6,    text width=0.70\textwidth},
  data/.style={base, draw=blue!50!black,  fill=blue!7,     text width=0.66\textwidth},
  lane/.style={base, draw=blue!50!black,  fill=blue!7,     text width=0.42\textwidth},
  vlm/.style ={base, draw=green!45!black, fill=green!9,    text width=0.42\textwidth},
  swin/.style={base, draw=orange!60!black,fill=orange!12,  text width=0.42\textwidth},
  eval/.style={base, draw=violet!55!black,fill=violet!9,   text width=0.70\textwidth},
  elab/.style={font=\footnotesize\itshape, fill=white, inner sep=1.5pt},
]
  \node[io] (cfg) {\textbf{Configuration} --- simulateSearch, global seed~42, L band 1000--1500\,MHz, 2048 channels, 8\,bit};
  \node[data, below=of cfg] (sim) {\textbf{Simulation} $\rightarrow$ PSRFITS \,+\, \texttt{labels.jsonl}\\[2pt]3000 samples simulated, 1000 per class\\[2pt]binary benchmark: balanced subset of 2000 --- 1000~\texttt{FRB} \;\textbar\; 500~\texttt{RFI} (5 subtypes) \;\textbar\; 500~\texttt{NOISE}};
  \node[lane, below=14mm of sim, xshift=-0.24\textwidth] (img) {\textbf{Dynamic-spectrum imaging} --- percentile norm., \texttt{viridis}, $1024\times768$\\[2pt]anonymized PNGs (\texttt{sample\_NNNNNN}) \,+\, \texttt{image\_manifest.jsonl}};
  \node[vlm,  below=12mm of img] (vlm)  {\textbf{Generalist VLM} (zero-shot, prompt-only)\\[2pt]Gemma 4 E2B / E4B, greedy decoding\\$\rightarrow$ structured JSON, \texttt{frb\_probability}};
  \node[swin] (swin) at ($(img)!0.5!(vlm)+(0.48\textwidth,0)$) {\textbf{Specialized baseline} (SwinYNet)\\[2pt]dedicated FRB detector, operates on FITS\\$\rightarrow$ continuous FRB score};
  \node[eval, below=24mm of vlm, xshift=0.24\textwidth] (eval) {\textbf{Paired evaluation by \texttt{sample\_id}}\\[2pt]accuracy / macro-F1 \,\textbullet\, ROC-AUC / AP \,\textbullet\, McNemar \,\textbullet\, false positives by RFI / NOISE};
  \node[io, below=of eval] (res) {\textbf{Results} --- metric tables, ROC curves, score histograms};

  \draw[flow] (cfg) -- (sim);
  \coordinate (f) at ($(sim.south)+(0,-7mm)$);
  \draw[link] (sim.south) -- (f);
  \draw[flow] (f) -| (img.north);
  \draw[flow] (f) -| (swin.north);
  \draw[flow] (img.south) -- (vlm.north);
  \node[elab, above=1.5pt of img]  {PSRFITS};
  \node[elab, above=1.5pt of swin] {PSRFITS};
  \node[elab, above=1.5pt of vlm]  {PNG + prompt};
  \coordinate (m) at ($(eval.north)+(0,9mm)$);
  \draw[link] (vlm.south)  |- (m);
  \draw[link] (swin.south) |- (m);
  \node[elab, below=1.5pt of vlm]  {threshold 0.5};
  \node[elab, below=1.5pt of swin] {threshold 0.5};
  \draw[flow] (m) -- (eval.north);
  \draw[flow] (eval) -- (res);
\end{tikzpicture}
\caption{\label{fig:pipeline}Overview of the end-to-end pipeline, from simulation to results. A controlled set of 3000 dynamic spectra (1000 per source class) is simulated with simulateSearch, producing native PSRFITS files; the binary benchmark shown here is a balanced subset of 2000 samples (1000~\texttt{FRB}, 500~\texttt{RFI} across five subtypes, and 500~\texttt{NOISE}). These PSRFITS then feed two independent detectors through different paths. For the generalist VLM (Gemma 4 E2B/E4B) they are first rendered as anonymized PNG images, which the model classifies from the image and a fixed prompt alone, whereas the specialized baseline SwinYNet operates directly on the native PSRFITS. Each produces a continuous FRB score thresholded at 0.5, and the two are then compared sample by sample through the shared \texttt{sample\_id}. The deliberate input asymmetry, a rendered image for the VLM versus the native FITS for SwinYNet, is part of the experimental design.}
\end{figure*}

This section describes the experimental design used to build the set of dynamic spectra evaluated in this work. The goal is not to reproduce the full complexity surrounding real observations.  The idea is to create a controlled, balanced, and reproducible benchmark to compare generalist VLMs with a specialized detector on the same samples. For this reason, the instrumental configuration was chosen to approximate the conditions used in the evaluation of SwinYNet, making the comparison more informative without presenting the set as a real observational validation. An overview of the complete pipeline, from simulation to the paired comparison of results, is shown in Fig.~\ref{fig:pipeline}.

\subsection{Motivation for using simulated data}

The use of synthetic data is a deliberate methodological choice in this first experiment. In real observational data, labels can be ambiguous, incomplete, or dependent on later decisions of human inspection. In a study that seeks to measure whether a generalist VLM recognizes visual patterns associated with FRBs, RFI, and noise, this ambiguity would make it difficult to separate model error from uncertainty in the label itself.

With simulation, each sample has a known ground truth. For the FRB examples, the physical parameters of the injected event are recorded in the metadata, including arrival time, DM, width, flux density, reference frequency, and profile type. For the RFI examples, the interference category and the parameters used in the generation are also preserved. For noise, the sample is defined as the absence of an injected coherent structure, containing only system noise.

Simulation also makes it possible to control the statistical composition of the benchmark. The benchmark for the binary task was built in a balanced way, containing the same number of \texttt{FRB} and \texttt{NON\_FRB} samples. This prevents the initial metrics from being dominated by an arbitrary class prevalence and makes the comparison between the VLM and the specialized detector more direct. The real prevalence of FRBs in astronomical surveys is much lower, but the goal of this stage is to evaluate the discriminative capability in a controlled environment.

Another important advantage is reproducibility. The pipeline uses a global seed and per-sample derived seeds. The same set can therefore be regenerated by other researchers as long as they use the same configuration, the same code version, and an equivalent environment for simulateSearch. This property is especially relevant because the article compares models sample by sample, not only by aggregate metrics.

\subsection{Generation of the PSRFITS files}

The data were generated with simulateSearch \cite{Luo2022_simulateSearch_paper, hobbs2022simulatesearch}, a package for simulating high time-resolution radio observations. The software allows one to compose system noise, dispersed transient signals, and different forms of RFI, producing files in search-mode PSRFITS format~\cite{hotan2004psrchive}. These files can be processed by conventional astronomical tools and, in this work, also serve as the source for the dynamic-spectrum images supplied to the VLMs.

Each simulated file corresponds to 2 s of observation. The configuration used has 2048 frequency channels, a time sampling of $1.96608\times10^{-4}$~s, and 8-bit quantization. In practice, each sample contains approximately 10\,173 time samples. The system noise level was varied reproducibly per sample through a dimensionless scale applied to the 25 K base $T_{\mathrm{sys}}$, yielding an effective system temperature between 16.25 K and 36.25 K and preventing the global noise level from being a fixed class cue.

Table~\ref{tab:instr} summarizes the main instrumental parameters used in the generation.

\begin{table*}
\caption{\label{tab:instr}Main instrumental parameters used in the generation of the simulated files.}
\begin{ruledtabular}
\begin{tabular}{lc}
Parameter & Value \\
\hline
Instrumental reference & aligned to SwinYNet domain \\
Start frequency & 1000 MHz \\
End frequency & 1500 MHz \\
Number of channels & 2048 \\
Time sampling & $1.96608\times10^{-4}$ s \\
Duration per file & 2 s \\
Time samples per file & $\approx 10\,173$ \\
Quantization & 8 bits \\
Gain & 0.7 \\
Base system temperature & 25 K \\
$T_{\mathrm{sys}}$ scale per sample (dimensionless) & 0.650187 to 1.449952 \\
Effective system temperature & 16.25 K to 36.25 K \\
Maximum FRB width & 0.005 s \\
Global seed & 42 \\
\end{tabular}
\end{ruledtabular}
\end{table*}

\subsection{Benchmark composition}

The simulation produced a full set of 3000 samples, with 1000 examples of each source class (\texttt{FRB}, \texttt{RFI}, and \texttt{NOISE}). The benchmark used in the binary evaluation is a balanced subset of 2000 of these samples, drawn deterministically from the full set with a fixed selection seed and summarized in Table~\ref{tab:benchmark}. The positive class is composed of the 1000 \texttt{FRB} examples. The negative class \texttt{NON\_FRB} contains 1000 examples, split equally between 500 \texttt{RFI} examples and 500 \texttt{NOISE} examples. The remaining 500 \texttt{RFI} and 500 \texttt{NOISE} samples are not discarded; the full set of 3000 is used in the multiclass task (Sect.~\ref{sec:multiclass-results}). The original labels are preserved in the metadata to allow diagnostic analyses, but the main task of the article aggregates \texttt{RFI} and \texttt{NOISE} into \texttt{NON\_FRB}, since this is the most direct formulation for comparison with a binary FRB detector.

\begin{table}
\caption{\label{tab:benchmark}Composition and class mapping of the binary benchmark data set. The 2000 total samples are evenly distributed between the target class (\texttt{FRB}) and the negative class (\texttt{NON\_FRB}), which combines structured interference (\texttt{RFI}) and background \texttt{NOISE}.}
\begin{ruledtabular}
\begin{tabular}{llr}
Source class & Binary label & Number of samples \\
\hline
\texttt{FRB} & \texttt{FRB} & 1000 \\
\texttt{RFI} & \texttt{NON\_FRB} & 500 \\
\texttt{NOISE} & \texttt{NON\_FRB} & 500 \\
\hline
\textbf{Total} & & \textbf{2000} \\
\end{tabular}
\end{ruledtabular}
\end{table}

This separation between the binary label and the source class is important for interpreting the results. Two samples incorrectly classified as FRB may have very different physical or instrumental causes. One may be rare stochastic noise, while another may be structured RFI. Preserving the source class makes it possible to estimate false-positive rates separately for RFI and noise, which became one of the central analyses of the article.

\subsection{Parameters of the simulated FRBs}\label{sec:frbparams}

For the \texttt{FRB} class, each sample receives a dispersed burst with reproducibly drawn parameters. In the metadata of the current set, the DM ranges from approximately 120.70 to 899.14~pc\,cm$^{-3}$, the temporal width ranges from 0.003 to 0.005 s, and the flux density ranges from approximately 4.02 to 11.98 in the internal units used by the simulation. The arrival time, the reference frequency, the dispersion index, and the profile type also vary.

These variations are relevant because they prevent the VLM from memorizing a single FRB morphology. Instead, the model must recognize a family of visual patterns associated with dispersed transients, such as temporal localization, broadband structure, and a systematic delay with frequency. Since the study does not estimate the DM directly, it enters here as a generation parameter and as a diagnostic variable for future work, not as a prediction target in this first article.

\subsection{RFI and noise}

The \texttt{RFI} examples were simulated to represent structured interference capable of confusing an automatic transient search. The full simulated set contains five categories, balanced with 200 samples each, of which 100 per category enter the binary benchmark (Table~\ref{tab:rfi}).

\begin{table}
\caption{\label{tab:rfi}Distribution of the 500 RFI samples of the binary benchmark across distinct interference categories. The benchmark includes five balanced categories (100 samples each) representing common environmental and anthropogenic interference types, ranging from continuous narrow-band signals to transient broad-band events.}
\begin{ruledtabular}
\begin{tabular}{lr}
RFI category & Number of samples \\
\hline
\texttt{persistent\_narrowband} & 100 \\
\texttt{impulsive\_broadband} & 100 \\
\texttt{wifi\_multiband} & 100 \\
\texttt{point\_to\_point\_microwave} & 100 \\
\texttt{satellite\_like\_inband} & 100 \\
\end{tabular}
\end{ruledtabular}
\end{table}

These categories cover both persistent contamination in frequency and impulsive or multiband events. This diversity is important because RFI is not a single visual class. Some samples may appear as persistent horizontal lines, others as broadband impulses, multiband structures, or patterns that occupy specific regions of the band. A useful classifier must separate these morphologies from a dispersed track compatible with an FRB.

The \texttt{NOISE} examples, in turn, represent samples without an injected coherent structure. They are important to test whether the model confuses random fluctuations with FRB candidates. The reproducible variation in the system noise level is applied to all classes, including FRB, RFI, and NOISE, to reduce the risk that the classifier uses only the global brightness or the mean contrast as a decision cue.

\subsection{Conversion to dynamic-spectrum images}

\begin{figure*}
\includegraphics[width=\textwidth]{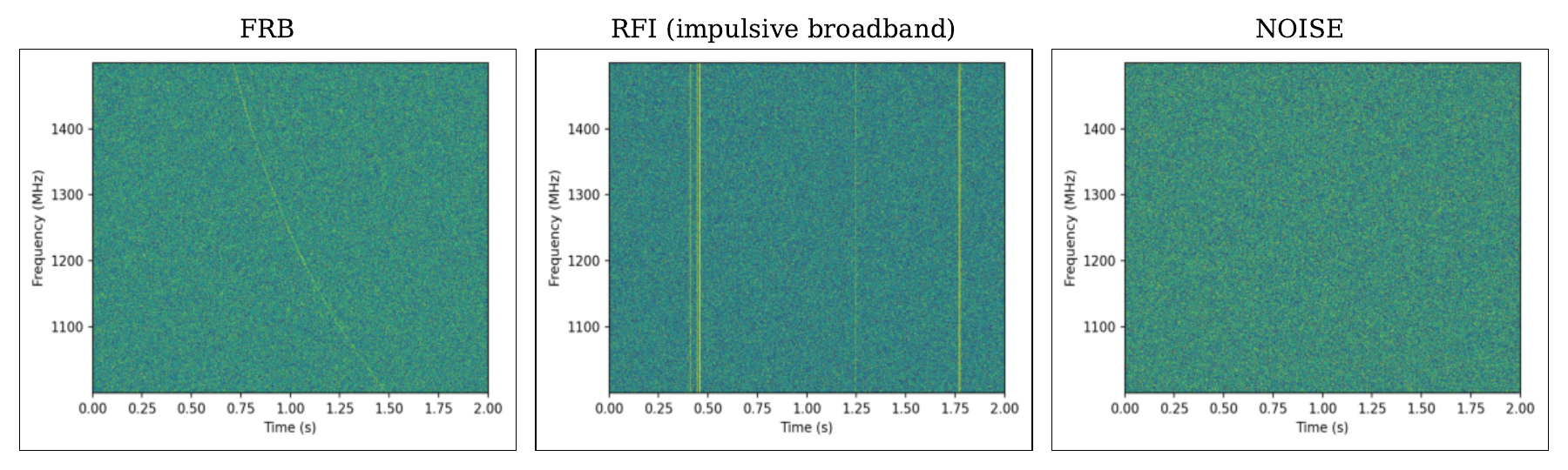}
\caption{\label{fig:examples}Representative anonymized dynamic-spectrum images, one per class, as supplied to the VLM. Left: an \texttt{FRB} (a single dispersed broadband transient); middle: \texttt{RFI} of the \texttt{impulsive\_broadband} type (broadband impulses localized in time); right: \texttt{NOISE} (system noise only). In each panel the horizontal axis is time, the vertical axis is frequency, and the pixel intensity is signal power (\texttt{viridis} colormap, percentile normalization).}
\end{figure*}

After the PSRFITS were generated, each file was converted into a PNG dynamic-spectrum image. In these images, the horizontal axis represents time, the vertical axis represents frequency, and the pixel intensity represents signal power. This representation was chosen because it corresponds to the type of visual input that a VLM can process directly. One representative example of each class is shown in Fig.~\ref{fig:examples}.

In the experimental run described here, the images were generated with percentile normalization, in which the pixel intensities are linearly scaled between the 1st and 99th percentiles of each dynamic spectrum and clipped to this range, with the \texttt{viridis} colormap, a width of 1024 pixels, and a height of 768 pixels. These parameters were recorded in the image manifest to keep the evaluation auditable. The choice of normalization and colormap can influence the performance of visual models; for this reason, it must be treated as part of the experimental protocol, not as a purely aesthetic detail.

The VLM receives only the final image and the textual classification instruction. It does not receive the FITS file, the simulation parameters, the DM, the true label, or any numerical metadata. This choice creates an important difference relative to SwinYNet, which operates on FITS files and can exploit a specialized pipeline. The comparison is therefore deliberately asymmetric. The specialized detector receives an input closer to its natural domain, while the VLM operates on a single visual representation.

\subsection{Anonymization and leakage control}

A central methodological point is to prevent label leakage. If the file name, the directory, or the image title contained terms such as \texttt{FRB}, \texttt{RFI}, or \texttt{NOISE}, a VLM could exploit that textual cue instead of analyzing the dynamic spectrum. To avoid this, the images supplied to the model were saved in a neutral directory, with anonymous names in the format \texttt{sample\_000000.png}, \texttt{sample\_000001.png}, and so on.

The image manifest preserves the link between the anonymous identifier, the source FITS, the simulation parameters, and the true label. This manifest is used only by the evaluation pipeline. The anonymous image path is the only path needed for VLM inference, and the images have no title with class information. In this way, the prediction must depend on the visual content of the dynamic spectrum.

This leakage control is also important for external comparison. The files and manifests make it possible to align predictions by \texttt{sample\_id} without exposing the source class to the classifier. The aggregate metrics and the paired comparison can thus be computed later without contaminating the classification stage.

\subsection{Reproducibility and generated artifacts}

The pipeline design explicitly separates data, images, predictions, and metrics. The simulation generates the PSRFITS files and the label metadata; the plotting stage generates the anonymized images and the visual manifest; the classification produces per-sample predictions; and the evaluation computes aggregate metrics and paired analyses. This separation makes it possible to repeat a single stage, audit inputs and outputs, and compare different models on exactly the same samples.

The main artifacts used in this stage are:
\begin{itemize}
\item \texttt{dataset/fits/}: simulated PSRFITS files;
\item \texttt{dataset/images/samples/}: anonymized images supplied to the VLM;
\item \texttt{dataset/metadata/labels.jsonl}: labels and simulation parameters;
\item \texttt{dataset/metadata/image\_manifest.jsonl}: link between the anonymous image, the source FITS, and the true label.
\end{itemize}

This set of artifacts defines the benchmark used in the following sections. From it, Sect.~\ref{sec:vlm} describes how the images were presented to the VLMs, how the responses were structured, and how the predictions were aligned with the outputs of the specialized detector.

\section{Using a VLM for FRB detection}\label{sec:vlm}

This section describes how the VLMs were used to classify the images generated in Sect.~\ref{sec:simulation}. The central point of the protocol is that the models were not trained, fine-tuned, or specialized for radio astronomy. They receive only the anonymous dynamic-spectrum image and a fixed textual instruction. The experiment thus tests a specific question. To what extent can a generalist VLM, used in a prompt-only regime, recognize visual patterns associated with FRBs and separate them from RFI and noise?

The main task of the article is binary, \texttt{FRB} versus \texttt{NON\_FRB}, to allow a direct comparison with the specialized detector. The multiclass formulation, \texttt{FRB}/\texttt{RFI}/\texttt{NOISE}, reveals a natural advantage of VLMs and reinforces the potential thesis, by showing that the task can be redefined through the prompt without changing the model weights.

\subsection{Input supplied to the model}

Each sample sent to the VLM consists of an anonymous PNG dynamic-spectrum image. The image is accompanied by a textual instruction, but no simulation metadata is provided to the model. In particular, the VLM does not receive the true label, the original FITS file name, the source class, the DM, the RFI parameters, or any identifier containing \texttt{FRB}, \texttt{RFI}, or \texttt{NOISE}.

The instruction informs only how to interpret the image visually, with the horizontal axis as time, the vertical axis as frequency, and the pixel intensity as signal power. The model must infer the class from the visual morphology of the dynamic spectrum. This restriction is important because it differentiates the VLM task from the task of the specialized detector. The VLM operates on a single image, while SwinYNet is evaluated in its natural domain, from FITS files and its own processing pipeline.

\subsection{Models evaluated}\label{sec:models}

Two models from the Gemma 4 family were evaluated (Table~\ref{tab:models}), namely the checkpoints \texttt{google/gemma-4-E2B-it} and \texttt{google/gemma-4-E4B-it}, referred to throughout the article as Gemma 4 E2B and Gemma 4 E4B. The \texttt{E} in the official designation denotes effective parameters, the count that governs memory and compute at inference; the raw parameter total of each checkpoint is larger, so the labels should not be read as plain parameter counts. Both are generalist, open-weight multimodal models that can be run locally. In the protocol of this work, they are used strictly in zero-shot mode, that is, no weight is updated, no FRB-specific training set is used, and no labeled example is included in the prompt as a demonstration.

This choice is deliberate. A specialized supervised classifier, such as SwinYNet, is expected to be a strong reference because it was designed for FRB searches. The VLM, on the other hand, is tested as a generalist model. Therefore, the result of interest is not only the final metric, but the capacity for visual transfer. Can the model apply general visual and textual knowledge to a specialized scientific domain without fine-tuning?

Local execution is also part of the motivation. Small models reduce the dependence on commercial APIs, per-token costs, and remote infrastructure. However, the operational evaluation must be made with caution. In the measured runs, the complete classification corresponds to a mean wall-clock time of 2.7 s per file for Gemma 4 E2B and 4.9 s per file for Gemma 4 E4B, while each simulated file contains 2 s of observation. These values indicate feasibility for candidate triage or hybrid pipelines, but they do not, by themselves, demonstrate a brute-force real-time search; the real-time margin of the lighter multiclass configuration is discussed in Sect.~\ref{sec:cost}.

\begin{table*}
\caption{\label{tab:models}Vision-Language Models evaluated and their role in the study.}
\begin{ruledtabular}
\begin{tabular}{lllr}
Model & Role in the study & Usage mode & Mean wall-clock time per file (binary task) \\
\hline
Gemma 4 E2B & Smaller generalist VLM & zero-shot, prompt-only & 2.7 s \\
Gemma 4 E4B & Larger generalist VLM & zero-shot, prompt-only & 4.9 s \\
\end{tabular}
\end{ruledtabular}
\end{table*}

The timing measurements above were obtained on a workstation with an Intel Xeon w9-3495X processor (56 cores / 112 threads, up to 4.8 GHz), about 1 TB of RAM, and 4 NVIDIA RTX 4500 Ada Generation GPUs (24 GB each). Inference was run with the Hugging Face Transformers library~\cite{wolf2020transformers}, using \texttt{device\_map="auto"}, the model's native precision, and greedy decoding, processing one image at a time, without batching. The mean times are computed as the total wall-clock duration of each complete classification run divided by the number of files, and therefore include image loading, preprocessing, and output parsing. It is worth noting that the models themselves are small enough to fit on a single 24 GB GPU and were designed for execution on consumer hardware: Gemma 4 E2B targets mobile devices and Gemma 4 E4B targets personal computers.\footnote{\url{https://ai.google.dev/gemma/docs/core}} The reported times therefore reflect this specific configuration, in a sample-by-sample regime and without optimization, and should be read as indicative, not as the latency expected on a typical phone or personal computer.

\subsection{Binary prompt and structured output}

In the binary task, the prompt asks the model to classify each image as \texttt{FRB} or \texttt{NON\_FRB}. The instruction defines \texttt{FRB} as a single localized transient, plausibly broadband, with a frequency-dependent arrival trend. The prompt also makes explicit that cases dominated by random noise, persistent bands, repetitive patterns, broadband impulses without a dispersion delay, saturation, blank regions, or normalization artifacts must be classified as \texttt{NON\_FRB}. The full binary prompt used is reproduced in Appendix~\ref{app:prompt}.

The expected response is a JSON object. The most important field for the evaluation is \texttt{frb\_probability}, a continuous estimate of the probability that the image contains an FRB. In addition, the model returns a textual label, a confidence in the discrete label, a short justification, and boolean visual attributes. The expected structure is shown in Listing~\ref{lst:json}.

\begin{lstlisting}[float=*,caption={Expected JSON structure returned by the model in the binary task.},label={lst:json}]
{
  "label": "FRB|NON_FRB",
  "frb_probability": 0.0,
  "confidence": 0.0,
  "reason": "short explanation referencing the visual evidence",
  "features": {
    "visible_structure": true,
    "broadband": true,
    "localized_transient": true,
    "frequency_dependent_delay": true,
    "persistent_bands": false,
    "repeating_pattern": false,
    "uniform_or_constant_background": false,
    "saturated_or_clipped": false,
    "random_background_texture": false
  }
}
\end{lstlisting}

It is worth clarifying how \texttt{frb\_probability} is obtained, since it does not come from a closed-form formula computed over the image pixels. The value is reported by the model itself in the corresponding JSON field, in response to the prompt instruction to estimate, in the interval [0,1], the probability that the image contains an FRB. The prompt provides a calibration guide that associates probability ranges with different degrees of visual evidence, from a clear, textbook broadband dispersed sweep (0.95--0.99) to an unambiguous artifact such as saturation or clipping (0.01--0.04), and is reproduced in full in Appendix~\ref{app:prompt}. The pipeline only validates and lightly normalizes this number, converting any percentages to the [0,1] scale and clamping to the valid interval, and then uses it directly as the continuous score in the ROC and Precision-Recall curves; the binary label comes from the same score through the 0.5 threshold, as described in Sect.~\ref{sec:threshold}. Because it is a number that the model chooses, and not a quantity derived from the internal token probabilities, in practice \texttt{frb\_probability} takes few distinct values, a quantization discussed in Sect.~\ref{sec:prob}.

Separating \texttt{frb\_probability} from \texttt{confidence} avoids a common ambiguity in evaluations with LLMs. The FRB probability represents the continuous score used for ROC curves, Precision-Recall curves, and calibration analysis. The confidence, on the other hand, represents only the degree of certainty declared by the model about the discrete label. For this reason, the probabilistic analyses of the article use \texttt{frb\_probability}, not the \texttt{confidence} field.

\subsection{Binary decision and operating threshold}\label{sec:threshold}

The binary label used in the metrics is derived from \texttt{frb\_probability} by a decision threshold. The default threshold is 0.5. Samples with \texttt{frb\_probability} $\ge 0.5$ are evaluated as \texttt{FRB}, and samples below this value are evaluated as \texttt{NON\_FRB}. This procedure ensures that the discrete decision is consistent with the continuous score used in the performance curves.

The textual label produced by the model is preserved, but it is not the primary source of the decision when a valid probability exists. If the model's text and the probability diverge, the sample receives a content warning for auditing, instead of being re-run until it produces a more convenient response. This methodological decision avoids rejection sampling and makes the evaluation more honest.

\subsection{Deterministic decoding}\label{sec:decoding}

All VLM inferences were made with deterministic decoding, in greedy mode, without sampling. In practice, this corresponds to the use of temperature 0. Parameters such as \texttt{top\_p} and \texttt{top\_k} are not activated in this regime. The goal is to reduce the stochastic variation between runs, since the same image, the same prompt, and the same model should produce the same response, within the limitations of the execution environment. This choice responds, in part, to the instability observed in generalist VLMs, whose outputs can vary with factors such as the decoding temperature~\cite{drozdova2025radio}. By fixing deterministic decoding, we eliminate this stochastic component of the variation, although not the sensitivity to the prompt design itself, taken up again in Sect.~\ref{sec:threats}.

This choice is important for reproducibility. If sampling were used, part of the differences between models or runs could reflect decoding variation, not visual capability. Since the study compares the VLMs with a specialized detector in a paired evaluation, reducing this source of variance makes the interpretation of the disagreements clearer.

\subsection{Handling invalid responses and execution failures}

The protocol separates execution failures from content failures. Transient errors, such as input/output problems, timeouts, or recoverable memory failures, may allow retries. Invalid content responses, such as malformed JSON or the absence of \texttt{frb\_probability}, are not resampled. The raw response is preserved, the sample is flagged with a warning, and the evaluation records the occurrence.

This distinction is essential in order not to inflate performance artificially. If the model could be called repeatedly until it produced a valid and coherent JSON, the evaluation would also measure a response-selection strategy, not just the model's first decision. In the results reported for the binary task, all 2000 samples were evaluated with no samples discarded due to error or invalid response.

\subsection{Multiclass task by prompt}\label{sec:multiclass-task}

Besides the main binary task, the same pipeline allows a multiclass formulation with the three labels \texttt{FRB}, \texttt{RFI}, and \texttt{NOISE}. In this mode, the prompt defines the dispersed burst and pure stochastic noise explicitly, and treats \texttt{RFI} as a residual class that collects any coherent structure that is neither of the two. The multiclass task is not used as the main axis of comparison with SwinYNet, because the specialized baseline is analyzed here in the binary formulation. Even so, the multiclass task is indeed relevant.

The interest of the multiclass task is to show that the role of the VLM can go beyond an \texttt{FRB} versus \texttt{NON\_FRB} decision. On receiving a different instruction, the model can attempt to separate false positives associated with RFI from samples dominated by pure noise. This flexibility is an important property of generalist models, since the task can be redefined through natural language, without retraining and without changing the architecture.

In this work, this flexibility is treated as supporting evidence of the potential of the VLMs, not as the central axis of the evaluation. The main quantitative comparison remains the paired binary analysis against SwinYNet, and the multiclass task does not compete with it; it adds a demonstration that the same tool, without retraining, still distinguishes \texttt{FRB}, \texttt{RFI}, and \texttt{NOISE} when the prompt asks for three classes.

\subsection{Comparison with the specialized detector}\label{sec:comparison}

The VLM predictions are compared with the SwinYNet predictions by anonymous sample identifier. The comparison is paired: each \texttt{sample\_id} has a true label, a VLM prediction, and a prediction from the specialized detector. This alignment makes it possible to measure not only aggregate metrics, but also agreements and disagreements sample by sample.

The specialized baseline was used exactly as released by its authors, with no retraining, fine-tuning, or architectural modification. We ran SwinYNet version 1.0.0, obtained from the official repository,\footnote{\url{https://github.com/expnn/SwinYNet}} with the pretrained weights distributed by the authors and the inference configuration provided in the released implementation. Inference was applied to anonymized copies of the simulated PSRFITS files, carrying the same neutral \texttt{sample\_NNNNNN} identifiers used for the images, so that the leakage control described in Sect.~\ref{sec:simulation} also holds for the specialized detector. Each detection reported by the model carries a confidence value, which is taken as its continuous FRB score.

For the binary comparison, the external SwinYNet outputs are normalized to the same representation used by the evaluation pipeline, with a continuous FRB score and a binary decision at the 0.5 threshold. When an external model produces more than one detection for the same sample, the maximum score across the detections is used as the sample-level probability before the metrics are computed.

It is important to stress again the input asymmetry between the approaches. SwinYNet operates on the FITS files and was developed specifically for FRB searches, including steps that can exploit physical and temporal information in more detail. The VLM receives only an anonymous PNG of the dynamic spectrum and a textual instruction. The comparison should therefore not be interpreted as a direct competition between equivalent architectures, but as a measure of the potential of a generalist approach against a reference specialized detector.

In the following results section, this paired structure is the basis to answer three questions: which model has the best aggregate performance at the default threshold?; how do the models behave as probabilistic classifiers?; and in which types of \texttt{NON\_FRB}, especially RFI or noise, each approach tends to produce false positives?

\subsection{Evaluation metrics}\label{sec:metrics}

To answer these three questions, the predictions of the two VLMs and of SwinYNet are evaluated on exactly the same 2000 samples, taking \texttt{FRB} as the positive class. The metrics used throughout the paper are defined below and apply to all the tables in the following sections, grouped into three sets: threshold-dependent metrics, probabilistic (threshold-independent) metrics, and statistical tests for the paired comparison.

\textit{Threshold-dependent metrics.} Once the decision threshold is fixed, each binary prediction is a true positive (TP), a false positive (FP), a false negative (FN), or a true negative (TN). From these counts, the accuracy is the fraction of correctly classified samples,
\begin{equation}
\mathrm{accuracy}=\frac{\mathrm{TP}+\mathrm{TN}}{\mathrm{TP}+\mathrm{TN}+\mathrm{FP}+\mathrm{FN}}.
\end{equation}
The precision of a class is the fraction of correct predictions among the samples predicted in that class, and the recall is the fraction of the samples of that class that the model recovered,
\begin{equation}
P=\frac{\mathrm{TP}}{\mathrm{TP}+\mathrm{FP}},\qquad
R=\frac{\mathrm{TP}}{\mathrm{TP}+\mathrm{FN}}.
\end{equation}
The F1 score is the harmonic mean of precision and recall, and the macro-F1 is the unweighted mean of the per-class F1 scores,
\begin{equation}
F_{1}=\frac{2PR}{P+R}.
\end{equation}
The false-positive rate (FPR) is the fraction of negative samples classified as positive,
\begin{equation}
\mathrm{FPR}=\frac{\mathrm{FP}}{\mathrm{FP}+\mathrm{TN}},
\end{equation}
also reported separately for RFI ($\mathrm{FPR}_{\mathrm{RFI}}$) and for noise ($\mathrm{FPR}_{\mathrm{NOISE}}$), using only the samples of each type. The specificity is its complement,
\begin{equation}
\mathrm{specificity}=\frac{\mathrm{TN}}{\mathrm{TN}+\mathrm{FP}}=1-\mathrm{FPR},
\end{equation}
and is therefore the same quantity as the recall of the \texttt{NON\_FRB} class. Both names are current in the literature, and both are used here, Table~\ref{tab:binary} reporting it as a per-class recall and Table~\ref{tab:realbinary} as a specificity.

\textit{Probabilistic metrics.} Independent of the threshold, these metrics evaluate the continuous \texttt{frb\_probability} score directly. The ROC-AUC is the area under the ROC curve and equals the probability that the model assigns a higher score to an \texttt{FRB} than to a \texttt{NON\_FRB} drawn at random, so that 1.0 indicates a perfect ranking and 0.5 indicates no discrimination. The Average Precision (AP) summarizes the Precision-Recall curve through the precision averaged over the recall gains. The Brier score~\cite{brier1950verification} is the mean squared error between the predicted probability and the true label, and the ECE (Expected Calibration Error)~\cite{guo2017calibration} measures the mean deviation between the predicted confidence and the observed accuracy; for both, lower values indicate better calibration.

\textit{Uncertainty and statistical tests.} A metric that is a proportion of a countable set, such as a recall, a specificity, or a precision, admits a 95\% Wilson interval~\cite{wilson1927probable}, which unlike the normal approximation stays inside $[0,1]$ and remains usable when the proportion approaches either bound. Metrics that combine two such proportions, such as the F1 scores and the accuracy, admit no such interval and are reported as point estimates. These intervals are given in Sect.~\ref{sec:real}, where several conclusions turn on whether two estimates are separated and where the strata are as small as a few dozen samples, and omitted in Sect.~\ref{sec:results}, where the differences under discussion are large compared with the sampling error on 2000 balanced samples. To compare two models on the same samples, the McNemar test~\cite{mcnemar1947note} checks whether the disagreements between them are statistically significant, with low p-values indicating a real difference at the considered threshold; we use its $\chi^{2}$ approximation with continuity correction for the full-benchmark comparison, and its exact binomial form when the test is restricted to a subset with strongly asymmetric discordant counts, as in the RFI-only analysis of Sect.~\ref{sec:fp}. The bootstrap confidence intervals~\cite{efron1979bootstrap} quantify the uncertainty of the differences (deltas) in ROC-AUC and AP between the models. Finally, in the multiclass task, the support of a class is the number of true samples of that class in the evaluated set.

\section{Results on the simulated benchmark}\label{sec:results}

This section gathers the quantitative results of the binary \texttt{FRB} versus \texttt{NON\_FRB} task. The comparison involves three systems evaluated on exactly the same 2000 balanced samples, namely the two generalist VLMs Gemma 4 E2B and Gemma 4 E4B, used only with a prompt, and the specialized detector SwinYNet, taken as a performance reference. The set contains 1000 \texttt{FRB} and 1000 \texttt{NON\_FRB}, with the negative class split into 500 \texttt{RFI} and 500 \texttt{NOISE}, and all samples were evaluated by the three models, with no record discarded due to inference error or invalid response.

The discussion follows the three questions raised at the end of Sect.~\ref{sec:comparison}. The aggregate performance at the default threshold is addressed in Sect.~\ref{sec:binary}, the behavior of the models as probabilistic classifiers in Sects.~\ref{sec:prob} and~\ref{sec:paired}, and the origin of the false positives in each type of \texttt{NON\_FRB} in Sect.~\ref{sec:fp}. Each result is discussed where it is presented, because the main metrics require immediate interpretation. The evaluation on real observations is deferred to Sect.~\ref{sec:real}.

\subsection{Binary performance at the default threshold}\label{sec:binary}

Table~\ref{tab:binary} summarizes the discrete metrics at the default operating threshold of 0.5. The central result of this section is that both generalist VLMs, without any task-specific training and operating only with a prompt, achieve strong performance, with accuracy between 0.90 and 0.94 in separating \texttt{FRB} from \texttt{NON\_FRB}. At the 0.5 operating point, Gemma 4 E2B recorded an accuracy of 0.9405 and a macro-F1 of 0.9404, values close to those of the specialized supervised detector SwinYNet (accuracy 0.9285, macro-F1 0.9281); Gemma 4 E4B obtained an accuracy of 0.9005 and a macro-F1 of 0.8999. As Sect.~\ref{sec:prob} makes clear, this proximity at the 0.5 threshold should not be read as superiority of the VLM over SwinYNet; the specialized detector has a perfect probabilistic ranking on this benchmark. It shows that a generalist model can reach performance similar to that of a dedicated detector starting from almost no domain information.

\begin{table*}
\caption{\label{tab:binary}Discrete metrics at the default operating threshold of 0.5, for the three systems, with precision and recall reported separately for \texttt{FRB} and for \texttt{NON\_FRB}. The metrics are defined in Sect.~\ref{sec:metrics}.}
\begin{ruledtabular}
\begin{tabular}{lcccccc}
Model & Accuracy & Macro-F1 & Precision (FRB) & Recall (FRB) & Precision (NON\_FRB) & Recall (NON\_FRB) \\
\hline
Gemma 4 E2B & 0.9405 & 0.9404 & 0.9742 & 0.9050 & 0.9113 & 0.9760 \\
Gemma 4 E4B & 0.9005 & 0.8999 & 0.9728 & 0.8240 & 0.8474 & 0.9770 \\
SwinYNet & 0.9285 & 0.9281 & 0.8749 & 1.0000 & 1.0000 & 0.8570 \\
\end{tabular}
\end{ruledtabular}
\end{table*}

The most important reading of the table is not a simple ranking of the models, but the error profile of each approach. SwinYNet recovered all the FRBs at the default threshold (\texttt{FRB} recall of 1.0000), that is, it produced no false negatives for the positive class. In contrast, it classified 143 \texttt{NON\_FRB} samples as \texttt{FRB}, reducing the precision of the positive class to 0.8749. Gemma 4 E2B was less complete for FRB (recall of 0.9050), but rejected the negative class better (\texttt{NON\_FRB} recall of 0.9760). Gemma 4 E4B was even more conservative, with high precision for FRB (0.9728), but lower FRB recall (0.8240).

This contrast of profiles is informative. SwinYNet behaves as a detector oriented toward high recall, minimizing FRB losses. The VLMs, especially Gemma 4 E2B, operate more selectively at the 0.5 threshold, rejecting the negative class better. Therefore, a generalist model that displays this selective behavior without specific training is already an indication of the potential of the approach, taken up again in Sect.~\ref{sec:fp}. The confusion matrices in Table~\ref{tab:confusion} reinforce this interpretation.

\begin{table}
\caption{\label{tab:confusion}Confusion matrices at the 0.5 threshold, with \texttt{FRB} as the positive class. TP, FN, FP, and TN are defined in Sect.~\ref{sec:metrics}.}
\begin{ruledtabular}
\begin{tabular}{lcccc}
Model & TP & FN & FP & TN \\
\hline
Gemma 4 E2B & 905 & 95 & 24 & 976 \\
Gemma 4 E4B & 824 & 176 & 23 & 977 \\
SwinYNet & 1000 & 0 & 143 & 857 \\
\end{tabular}
\end{ruledtabular}
\end{table}

In the table, TP represents correctly identified FRBs, FN represents missed FRBs, FP represents \texttt{NON\_FRB} samples classified as FRB, and TN represents \texttt{NON\_FRB} samples correctly rejected.

A point that deserves explicit comment is that the larger VLM performs worse than the smaller one at this operating point (accuracy 0.9005 against 0.9405). The score distributions discussed in Sect.~\ref{sec:prob} indicate the origin of this inversion. Gemma 4 E4B assigns low \texttt{frb\_probability} values more often than Gemma 4 E2B (1153 against 1071 of the 2000 samples fall below the 0.5 threshold), so the fixed cut converts more faint FRBs into false negatives (176 against 95). The inversion therefore reflects a more conservative scoring profile interacting with the fixed threshold and with the coarse quantization of the self-reported score, rather than a generalized visual deficit of the larger model; in the multiclass task of Sect.~\ref{sec:multiclass-results}, Gemma 4 E4B is consistently the better of the two.

\subsection{Probabilistic performance}\label{sec:prob}

\begin{figure*}
\includegraphics[width=0.49\textwidth]{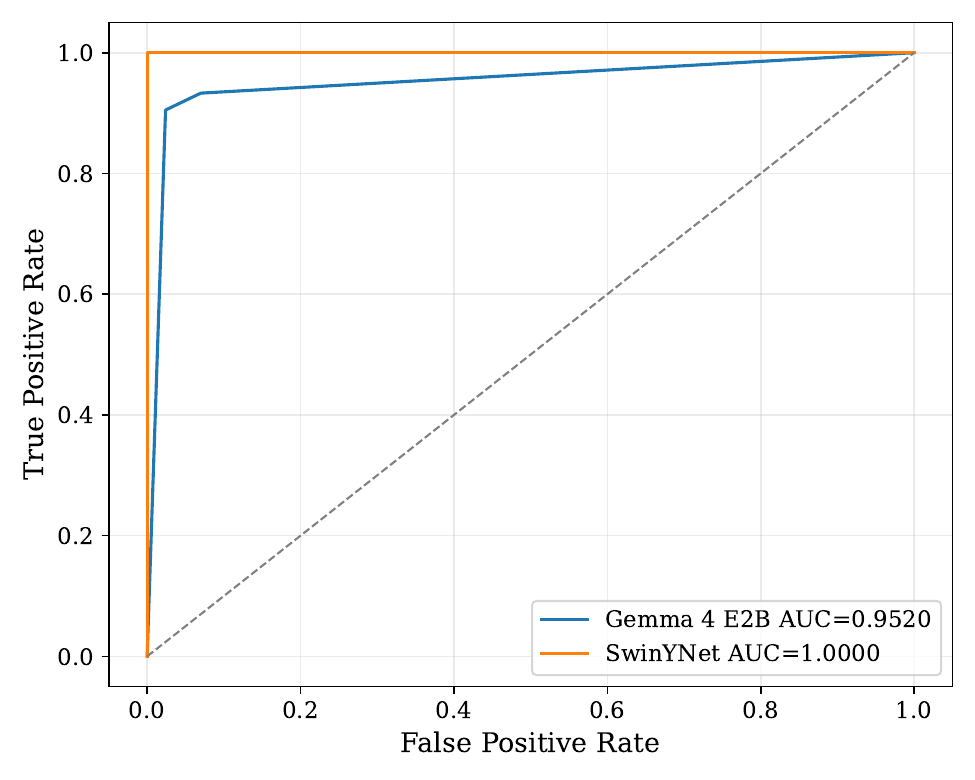}
\hfill
\includegraphics[width=0.49\textwidth]{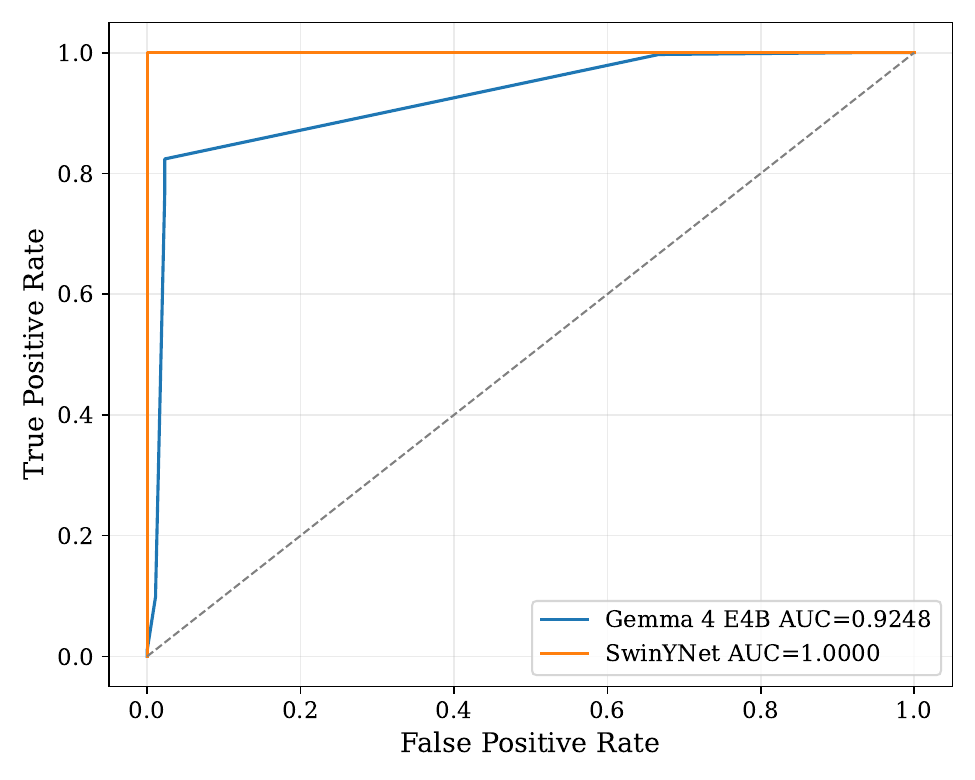}
\caption{\label{fig:roc}ROC curves for the binary \texttt{FRB} versus \texttt{NON\_FRB} task. Left: Gemma 4 E2B; right: Gemma 4 E4B. In each panel, the specialized detector SwinYNet attains a perfect ranking (AUC = 1.0000), whereas the staircase shape of the VLM curve reflects the small number of distinct \texttt{frb\_probability} values.}
\end{figure*}

The continuous metrics, reported in Table~\ref{tab:prob}, place the performance of the VLMs relative to the ceiling of the specialized detector. SwinYNet showed perfect probabilistic discrimination on this benchmark, with a ROC-AUC of 1.0000 and an Average Precision of 1.0000, a ceiling that is partly by construction, since the instrumental configuration was aligned to the detector's native domain (Sect.~\ref{sec:simulation}). In our run, every \texttt{FRB} sample received a score above the highest \texttt{NON\_FRB} score (the lowest \texttt{FRB} score, 0.999999, exceeds the highest \texttt{NON\_FRB} score, 0.999453), which separates the classes perfectly by ranking. The generalist VLMs fall below this ceiling, but close to it: Gemma 4 E2B obtained a ROC-AUC of 0.9520 and an AP of 0.9412, and Gemma 4 E4B obtained a ROC-AUC of 0.9248 and an AP of 0.8991, a strong probabilistic performance for models without specific training. The corresponding ROC curves are shown in Fig.~\ref{fig:roc}.

\begin{table*}
\caption{\label{tab:prob}Probabilistic performance metrics computed over the continuous \texttt{frb\_probability} score, for the three systems. The metrics are defined in Sect.~\ref{sec:metrics} (for the Brier score and the ECE, lower values are better).}
\begin{ruledtabular}
\begin{tabular}{lcccc}
Model & ROC-AUC & Average Precision & Brier score & ECE \\
\hline
Gemma 4 E2B & 0.9520 & 0.9412 & 0.0628 & 0.1037 \\
Gemma 4 E4B & 0.9248 & 0.8991 & 0.0908 & 0.0680 \\
SwinYNet & 1.0000 & 1.0000 & 0.0546 & 0.1042 \\
\end{tabular}
\end{ruledtabular}
\end{table*}

This contrast requires a careful reading, so as neither to exaggerate nor to underestimate the result. Since SwinYNet ranks the classes perfectly, at an optimal threshold it would classify without errors; the 143 false positives it shows in Sect.~\ref{sec:binary} are an effect of the fixed 0.5 threshold, not a limit of the model. Hence, the accuracy proximity between Gemma 4 E2B and SwinYNet at the 0.5 point is, in part, a threshold artifact, and not evidence that the VLM is superior. The defensible conclusion is the opposite, and more interesting for the thesis of this article. Even starting from a perfect ceiling of the dedicated detector, a zero-shot generalist VLM recovers most of this discriminative capability (ROC-AUC in the 0.92--0.95 range) without seeing a single labeled example.

\begin{figure*}
\includegraphics[width=0.49\textwidth]{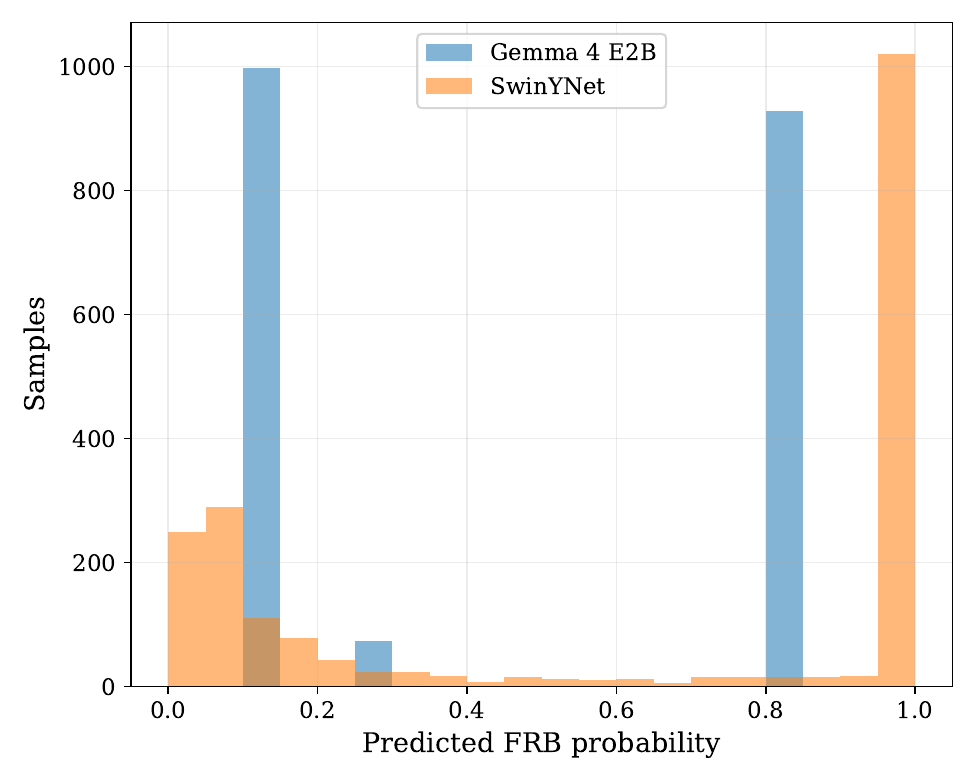}
\hfill
\includegraphics[width=0.49\textwidth]{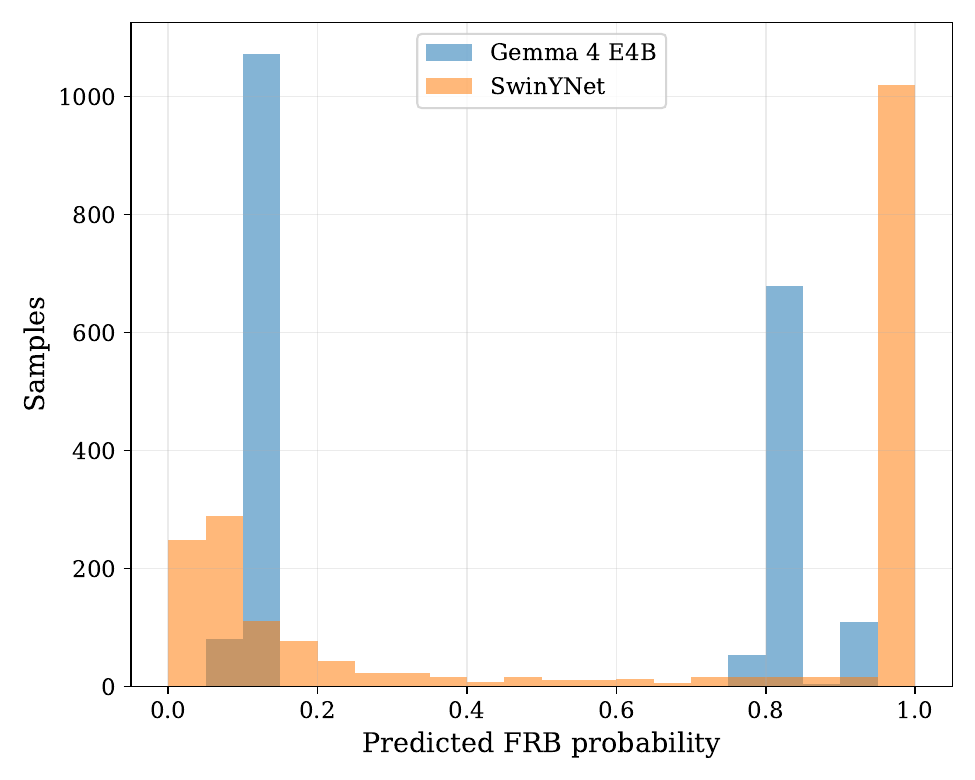}
\caption{\label{fig:scorehist}Histograms of the predicted \texttt{frb\_probability} for the binary task. Left: Gemma 4 E2B; right: Gemma 4 E4B. The VLM scores concentrate on a few discrete values, illustrating the strong quantization discussed in the text, whereas the specialized detector SwinYNet saturates near 0 and 1.}
\end{figure*}

It is also necessary to interpret these continuous metrics with caution, because of a limitation present in the results. The probabilities emitted by the VLMs are strongly quantized (Fig.~\ref{fig:scorehist}). Gemma 4 E2B used only three values of \texttt{frb\_probability} across the entire benchmark --- 0.15 (997 samples), 0.25 (74), and 0.85 (929) --- and Gemma 4 E4B used eight distinct values. In practice, \texttt{frb\_probability} currently functions more as an ordinal confidence level than as a continuous score, so that ROC-AUC, AP, ECE, and the Brier score derive from curves with very few operating points. SwinYNet, on the other hand, saturates at or extremely close to 1.0 for the positive samples; note that its perfect ranking does not translate into superior calibration (its ECE, 0.1042, is comparable to that of Gemma 4 E2B and worse than that of Gemma 4 E4B). For this reason, these numbers should be read as a diagnostic of the scoring behavior in this implementation, not as a definitive probabilistic calibration of the models.

\subsection{Paired comparison with SwinYNet}\label{sec:paired}

The sample-by-sample comparison, summarized in Table~\ref{tab:paired}, makes it possible to assess whether the differences at the 0.5 threshold are statistically evident and whether the models err on the same cases. Gemma 4 E2B and SwinYNet agreed on 1782 of the 2000 samples and disagreed on 218. Among the disagreements, Gemma 4 E2B was correct in 121 cases in which SwinYNet erred, while SwinYNet was correct in 97 cases in which Gemma 4 E2B erred. The McNemar test produced $p = 0.1193$, so the discrete difference between the two models at this operating point is not statistically significant under the approximation used.

Gemma 4 E4B showed a different profile. It disagreed with SwinYNet on 300 samples. It was correct in 122 cases in which SwinYNet erred, but SwinYNet was correct in 178 cases in which Gemma 4 E4B erred. In this case, the McNemar test produced $p = 0.0015$, indicating a significant advantage of SwinYNet at the 0.5 threshold.

\begin{table*}
\caption{\label{tab:paired}Paired comparison between each VLM and SwinYNet at the 0.5 threshold, over the same samples. The disagreements column counts the samples for which the two models assigned different labels, and the two following columns split them between the cases in which only the VLM was correct and those in which only SwinYNet was correct. The McNemar test reported in the last column is defined in Sect.~\ref{sec:metrics}.}
\begin{ruledtabular}
\begin{tabular}{lcccc}
Comparison & Disagreements & VLM right / SwinYNet wrong & SwinYNet right / VLM wrong & McNemar $p$ \\
\hline
Gemma 4 E2B vs SwinYNet & 218 & 121 & 97 & 0.1193 \\
Gemma 4 E4B vs SwinYNet & 300 & 122 & 178 & 0.0015 \\
\end{tabular}
\end{ruledtabular}
\end{table*}

The bootstrap analysis of the scores reinforces the probabilistic advantage of SwinYNet. For Gemma 4 E2B, the delta (VLM minus SwinYNet) was $-0.0480$ in ROC-AUC, with a 95\% confidence interval between $-0.0574$ and $-0.0380$; for Average Precision, the delta was $-0.0588$, with an interval between $-0.0719$ and $-0.0458$. For Gemma 4 E4B, the deltas were even more negative: $-0.0752$ in ROC-AUC and $-0.1009$ in Average Precision.

These results support a balanced reading that is favorable to the thesis of the VLMs' potential. At the default operating point, the discrete decisions of Gemma 4 E2B show no statistically significant difference from those of a dedicated detector (McNemar $p = 0.1193$), although SwinYNet remains superior as a probabilistic classifier on this benchmark. That a generalist model, without training and without access to the FITS, performs at the level of a specialized detector at the default threshold, with no statistically significant difference, is, by itself, strong evidence of what the approach is capable of doing.

\subsection{False positives by non-FRB type}\label{sec:fp}

The stratification of the false positives by source class is one of the most revealing results about what the generalist VLM actually learned. The \texttt{NON\_FRB} class aggregates RFI and noise, but these two groups represent different difficulties for a search pipeline. Structured RFI tends to produce more problematic false candidates than stochastic noise, because it can partially mimic coherent structures in the dynamic spectrum.

At the 0.5 threshold (Table~\ref{tab:fp}), SwinYNet classified 123 of the 500 RFI samples as FRB, producing $\mathrm{FPR}_{\mathrm{RFI}} = 0.246$. Gemma 4 E2B produced 24 false positives on RFI ($\mathrm{FPR}_{\mathrm{RFI}} = 0.048$), and Gemma 4 E4B produced 23 ($\mathrm{FPR}_{\mathrm{RFI}} = 0.046$). On noise, the VLMs produced no false positives, while SwinYNet classified 20 of the 500 noise samples as FRB ($\mathrm{FPR}_{\mathrm{NOISE}} = 0.040$). Because the three systems were evaluated on the same samples, this gap can be tested directly. Restricting the McNemar test to the 500 RFI samples gives $p = 1.1\times10^{-27}$ for Gemma 4 E2B versus SwinYNet (101 samples in which only the VLM is correct, against 2 in which only SwinYNet is correct) and $p = 5.4\times10^{-28}$ for Gemma 4 E4B (102 against 2), so the better RFI rejection of the VLMs at this operating point is statistically significant.

\begin{table*}
\caption{\label{tab:fp}False positives by \texttt{NON\_FRB} source type at the 0.5 threshold. The false-positive columns count, for each model, how many \texttt{RFI} and \texttt{NOISE} samples were classified as \texttt{FRB}, out of the 500 of each type. The FPR is defined in Sect.~\ref{sec:metrics}, with lower values indicating better rejection of that contaminant.}
\begin{ruledtabular}
\begin{tabular}{lcccc}
Model & False positives on RFI & FPR RFI & False positives on NOISE & FPR NOISE \\
\hline
Gemma 4 E2B & 24/500 & 0.048 & 0/500 & 0.000 \\
Gemma 4 E4B & 23/500 & 0.046 & 0/500 & 0.000 \\
SwinYNet & 123/500 & 0.246 & 20/500 & 0.040 \\
\end{tabular}
\end{ruledtabular}
\end{table*}

This table shows that the VLM is not merely recognizing the shape of an FRB. It actively rejects structured RFI, which is precisely the most difficult type of contaminant. At the default threshold, Gemma 4 E2B and E4B classified as FRB only 4.8\% and 4.6\% of the RFI samples, against 24.6\% for SwinYNet, and no VLM produced a false positive on pure noise. For a model that was never trained on this task, distinguishing structured interference from a dispersed track is a concrete demonstration of potential, and it suggests that the approach could, in real streams with many false RFI candidates, help reduce the inspection load.

This conclusion must be limited to the current benchmark. The data are simulated, the RFI categories are controlled, and the decision threshold was not optimized for a real survey. Even so, the result shows that the VLM goes beyond replicating the decision of the specialized detector; it errs differently, and this difference may have practical value.

\subsection{Computational cost and operational use}\label{sec:cost}

As detailed in Sect.~\ref{sec:models}, the mean wall-clock time of the binary task was 2.7 s per file for Gemma 4 E2B and 4.9 s per file for Gemma 4 E4B, measured on the workstation described there, in a sample-by-sample regime and without batching. Since each simulated file contains 2 s of observation, the binary configuration, taken in isolation, does not process these files faster than they are acquired.

The multiclass run of Sect.~\ref{sec:multiclass-results} shows that this margin is set by the prompt, not by the model. With the compact three-class instruction, the same models classify each file in 1.0 s (Gemma 4 E2B) and 1.5 s (Gemma 4 E4B), faster than the 2 s of observation each file contains. The difference tracks the amount of text processed per image. The binary instruction is about 1.5 times longer than the multiclass one, and the binary JSON response, with its probability field and per-feature block, averages about 520 characters against about 180 for the multiclass output, which matters because the response is generated serially, token by token. A leaner binary output is a prompt-only change that requires no retraining, and we measured its effect for both models on a stratified 200-image subsample of the benchmark (100 \texttt{FRB}, 50 \texttt{RFI}, 50 \texttt{NOISE}), running four prompt variants on the same images with a per-inference timer; the results are summarized in Table~\ref{tab:promptlatency}. A bare lean output that keeps only the label and the continuous probability cuts the generated text about twelvefold and the latency by a factor of about 7, and a lean variant that adds one short justification, generated before the decision fields, brings both models below the 2 s of observation while matching (Gemma 4 E2B) or exceeding (Gemma 4 E4B) the accuracy of the full prompt on the subsample. The two models differ in an instructive way. Gemma 4 E2B loses accuracy when the justification is removed (0.805 against 0.930), whereas Gemma 4 E4B does not (0.910 in both cases), so the brief reasoning generated before the decision is what sustains the sensitivity of the smaller model to faint FRBs. These controlled-run times are not directly comparable to the full-run figures above, since this run used a single GPU and the default dynamic cache; the internal comparison nevertheless shows that the operating point on the latency--accuracy plane is set by the prompt and its output format, and that faster-than-real-time binary configurations with no accuracy loss are already available for both models.

The same experiment answers a natural question, namely whether the main results should be reproduced with the faster prompts. The justified lean variant replicates the decisions of the full prompt on 97--98\% of the subsample labels at equal or better accuracy, so a full re-run is expected to change the full-benchmark results only within noise while changing their cost substantially; the bare lean variant of Gemma 4 E2B, in contrast, is flagged by the same protocol as non-equivalent (87.5\% agreement and lower accuracy). We therefore keep the full-prompt runs as the reference for classification quality and treat the paired subsample as an inexpensive audit of each prompt revision, reserving a full re-run for prompts whose subsample behavior departs from the reference.

\begin{table*}
\caption{\label{tab:promptlatency}Controlled prompt-latency run on a stratified 200-image subsample of the binary benchmark (100 \texttt{FRB}, 50 \texttt{RFI}, 50 \texttt{NOISE}), with four prompt variants applied to the same images on a single GPU and timed per inference. Chars is the mean length of the generated response, time is the mean wall-clock per file, the accuracy is computed at the 0.5 threshold on the subsample (binarizing the multiclass labels into \texttt{FRB} versus \texttt{NON\_FRB}), and the agreement is the fraction of labels identical to those of the full binary prompt. These times are internal to this run and not directly comparable to the full-run figures of Sect.~\ref{sec:models}.}
\begin{ruledtabular}
\begin{tabular}{llcccc}
Model & Prompt variant & Chars & Time (s) & Acc. & Agr. \\
\hline
Gemma 4 E2B & full binary & 524 & 4.06 & 0.930 & --- \\
 & lean & 43 & 0.56 & 0.805 & 0.875 \\
 & lean + reason & 182 & 1.16 & 0.930 & 0.980 \\
 & multiclass & 173 & 1.25 & 0.930 & --- \\
\hline
Gemma 4 E4B & full binary & 518 & 5.93 & 0.910 & --- \\
 & lean & 43 & 0.81 & 0.910 & 0.990 \\
 & lean + reason & 196 & 1.78 & 0.930 & 0.970 \\
 & multiclass & 180 & 1.69 & 0.945 & --- \\
\end{tabular}
\end{ruledtabular}
\end{table*}

For an end-to-end accounting, the VLM path also requires rendering the dynamic spectrum, which in our unoptimized implementation costs about 2.2 s per file, measured as the wall-clock duration of the complete 3000-image rendering pass. This stage is executed once per file, is CPU-bound, and is independent of the GPU inference, so in a pipelined implementation the sustained throughput is bounded by the slower of the two stages rather than by their sum; it is also trivially parallel across CPU cores and leaves ample room for optimization. For reference, the complete SwinYNet run on the same workstation, with the released inference configuration, corresponds to about 3.5 s per file end to end; we quote this number only as context, since neither pipeline was tuned for throughput.

These figures are specific to this benchmark, that is, simulated 2 s L-band files at the resolution of Table~\ref{tab:instr}, processed one at a time on the workstation of Sect.~\ref{sec:models}. They should be read as indicating that faster-than-real-time VLM triage is a realistic engineering target, not as a demonstration of real-time operation on a production stream. Within this reading, the measured cost is already perfectly acceptable for the triage of pre-selected candidates or for hybrid pipelines, and it tends to decrease further with batching, quantization, or even smaller models. It is also the scenario in which the textual explainability of the VLM adds the most value, since, in addition to deciding, the model justifies the decision in natural language, something rare in specialized detectors.

\subsection{Disagreements and cases exclusive to the VLM}

The disagreement files show that there are cases in which only the VLM is correct and cases in which only SwinYNet is correct. This information is richer than the aggregate accuracy. For Gemma 4 E2B, there were 121 samples in which the VLM was correct and SwinYNet erred, against 97 in which the opposite occurred. In addition, 22 samples were wrong for both models. For Gemma 4 E4B, there were 122 correct answers exclusive to the VLM, 178 correct answers exclusive to SwinYNet, and 21 common errors. The fact that there are dozens of samples that only the VLM gets right indicates that the generalist model captures visual patterns that the specialized detector lets through.

To characterize these disagreements, each sample was cross-referenced with its source class and simulation parameters, and the pattern is clear and almost symmetric.

The VLM's exclusive correct answers fall entirely on the negative class. For Gemma 4 E2B, the 121 samples in which the VLM is right and SwinYNet is wrong are 101 RFI and 20 noise, with no FRB, that is, cases in which the specialized detector produced a false positive and the VLM correctly rejected it. Among the RFI, the \texttt{wifi\_multiband} type dominates (75 of 101), followed by \texttt{impulsive\_broadband} (8), \texttt{point\_to\_point\_microwave} (8), \texttt{persistent\_narrowband} (5), and \texttt{satellite\_like\_inband} (5). The VLM's exclusive gain therefore lies precisely in structured RFI, especially multiband interference, the contaminant hardest to distinguish from a transient.

The exclusive errors of VLMs, in turn, are almost all faint FRBs. For Gemma 4 E2B, 95 of the 97 samples in which SwinYNet is right and the VLM is wrong are FRBs that the VLM let through (the other two are RFI). These FRBs are systematically of low intensity. The median flux density of the missed FRBs is 4.81, against 7.87 for the full FRB set (same internal units as Sect.~\ref{sec:frbparams}), and 67 of the 95 fall below the global 25th percentile (5.84); none exceeds 8.33, whereas the maximum of the set is 11.98. In the visual attributes reported by the model itself, these cases appear as having no visible structure and a random background texture. That is, the VLM perceived them as noise, consistent with FRBs of low visual intensity in the PNG.

Gemma 4 E4B repeats the same profile even more markedly. Of the 122 exclusive correct answers, 102 are RFI (76 \texttt{wifi\_multiband}) and 20 noise; and of the 178 exclusive errors, 176 are faint FRBs, with a median flux density of 5.37 and 104 of the 176 below the global 25th percentile (5.84).

This characterization turns the paired comparison into qualitative evidence of the VLM's behavior. The generalist model does not err randomly relative to the dedicated detector, since it gains exactly where SwinYNet stumbles, rejecting structured RFI that the detector confuses with an FRB, and loses exactly on the faintest FRBs, near the detection limit. In real streams with many false RFI candidates, this complementary behavior has practical value, although the loss of faint FRBs calls for caution regarding the fraction of FRBs recovered.

\subsection{Multiclass classification}\label{sec:multiclass-results}

In addition to the binary task, a multiclass formulation was tested with the three labels \texttt{FRB}, \texttt{RFI}, and \texttt{NOISE}. This experiment is not the main axis of the comparison with SwinYNet, but it illustrates the flexibility of the VLM in redefining the task through the prompt.

Both models were evaluated on the full simulated set of 3000 samples described in Sect.~\ref{sec:simulation}, with 1000 samples per class. This set is a superset of the binary benchmark, containing the same 1000 \texttt{FRB} samples together with all 1000 \texttt{RFI} examples (200 per subtype) and all 1000 \texttt{NOISE} examples, whereas the binary evaluation used 500 of each negative class. Gemma 4 E4B reaches an overall accuracy of 86.0\% (macro F1-score of 0.859) and Gemma 4 E2B reaches 80.1\% (macro F1-score of 0.799). The per-class metrics for both models are reported in Table~\ref{tab:multiclass}.

The error structure is shared by the two models and is informative in itself. The \texttt{FRB} class is recovered with perfect precision (1.0000). No \texttt{RFI} or \texttt{NOISE} sample is ever labeled \texttt{FRB}, so the reconfiguration to three classes introduces no false FRB. The \texttt{NOISE} class is recovered with perfect recall (1.0000). Every noise sample is identified as such. The whole difficulty concentrates in the \texttt{RFI} class, whose recall (0.6600 for Gemma 4 E4B and 0.5340 for Gemma 4 E2B) is limited by the faintest interference subtypes, which are systematically confused with \texttt{NOISE} rather than with \texttt{FRB}. Broken down by subtype, \texttt{persistent\_narrowband} and \texttt{wifi\_multiband} are recovered perfectly by both models, whereas the two faintest categories concentrate the loss. \texttt{point\_to\_point\_microwave} (recall 0.075 for Gemma 4 E2B and 0.260 for Gemma 4 E4B) and \texttt{satellite\_like\_inband} (0.075 and 0.325, in the same order) account for 370 of the 466 and 283 of the 340 \texttt{RFI}-to-\texttt{NOISE} errors of the two models, with \texttt{impulsive\_broadband} in an intermediate position (0.520 and 0.715). This mirrors the faint-FRB behavior of the binary task. An interference signal whose visual contrast approaches the noise floor of the rendered image is read as noise, in line with the dependence on the visual representation discussed in Sect.~\ref{sec:threats}.

\begin{table}
\caption{\label{tab:multiclass}Per-class metrics for the multiclass task (\texttt{FRB}/\texttt{RFI}/\texttt{NOISE}), for both VLMs, on 1000 samples per class. Metrics and support are defined in Sect.~\ref{sec:metrics}.}
\begin{ruledtabular}
\begin{tabular}{llcccc}
Model & Class & Precision & Recall & F1-score & Support \\
\hline
Gemma 4 E2B & \texttt{FRB} & 1.0000 & 0.8700 & 0.9305 & 1000 \\
 & \texttt{RFI} & 0.9889 & 0.5340 & 0.6935 & 1000 \\
 & \texttt{NOISE} & 0.6289 & 1.0000 & 0.7722 & 1000 \\
\hline
Gemma 4 E4B & \texttt{FRB} & 1.0000 & 0.9210 & 0.9589 & 1000 \\
 & \texttt{RFI} & 0.9565 & 0.6600 & 0.7811 & 1000 \\
 & \texttt{NOISE} & 0.7199 & 1.0000 & 0.8372 & 1000 \\
\end{tabular}
\end{ruledtabular}
\end{table}

The mean wall-clock time was 1.0 s per file for Gemma 4 E2B and 1.5 s per file for Gemma 4 E4B, well below the binary-task figures reported in Sect.~\ref{sec:models} and, notably, below the 2 s of observation contained in each file. In this configuration the models classify the simulated spectra faster than real time at the inference stage, as discussed in Sect.~\ref{sec:cost}. This reduction follows directly from the design of the multiclass prompt. The instruction is deliberately compact and defines \texttt{RFI} as a residual class, anything that is neither a single dispersed sweep nor pure noise, instead of enumerating the interference morphologies present in the benchmark, which would encode dataset-specific priors into the classifier. Its structured output is correspondingly lighter, dropping the probability field and the per-feature diagnostic block requested in the binary task, so that fewer instruction and output tokens are processed per image. Consistently with the demonstrative role of this experiment, no per-class prompt tuning was pursued; the multiclass prompt is reproduced in full in Appendix~\ref{app:multiclass-prompt}.

Even in this deliberately simple setting, the result is one of the clearest pieces of evidence of the potential pursued in this work. Without any retraining, and only by rewriting the prompt, the same model separates three classes and tells pure noise apart from structured RFI and from dispersed bursts. This reconfiguration of the task through natural language is something that a supervised detector does not offer without a new training cycle.

\section{Validation on real data}\label{sec:real}

The benchmark of Sect.~\ref{sec:results} is simulated, and a controlled set, however carefully built, does not capture the complexity of a real observation. This section addresses that limitation directly, applying the same two models, with the same prompt, the same operating threshold, and the same decoding configuration, to real observations from the Five-hundred-meter Aperture Spherical radio Telescope (FAST). Nothing in the classifier is retrained, re-tuned, or re-prompted for this experiment; the only change is the origin of the images.

\subsection{The FAST-FREX data set}\label{sec:frex}

The evaluation uses FAST-FREX (the FAST data set for Fast Radio bursts EXploration)~\cite{guo2025accelerating}, the public data set released alongside the RaSPDAM detector discussed in Sect.~\ref{sec:frb}. It contains 1600 PSRFITS files, each about 6.04 s long, recorded with 4096 frequency channels over 1000--1500 MHz, comprising 600 positive samples, each containing one catalogued burst, and 1000 negative samples of RFI and noise. The positives come from three repeating sources with a strongly uneven distribution, namely 470 bursts from FRB20121102, 125 from FRB20201124, and 5 from FRB20180301, and are accompanied by parameter tables giving, for each burst, its time of arrival within the file and its best-fit DM. The negatives carry no such parameters, since the data set does not separate interference from noise among them; both are therefore mapped to the \texttt{NON\_FRB} class, exactly as the \texttt{RFI} and \texttt{NOISE} classes of the simulated benchmark are.

The whole data set is evaluated. This is a deliberate choice, because with 600 positives the recall is determined to about $\pm 4$ percentage points, and the 1000 negatives make the false-positive rate measurable at the level of a few per mille. Unlike the simulated benchmark, this set is not balanced, with 37.5\% positives. Accuracy is therefore inflated by prevalence and is not comparable with the accuracy of Sect.~\ref{sec:binary}; recall, specificity, and macro F1-score are the quantities to be compared across the two experiments.

\subsection{From real PSRFITS to comparable images}\label{sec:realprep}

A real spectrum cannot be rendered in the same way as a simulated one. In the simulated benchmark the file \emph{is} the sample. It is exactly 2 s long, its bandpass is flat by construction, and it contains no dead channels or persistent interference, so the PSRFITS is decoded and passed directly to the image renderer. Real observations require the preprocessing that the simulation makes unnecessary, and this preprocessing is inserted before, and only before, the rendering step; the visual protocol itself (percentile 1--99 normalization, \texttt{viridis} colormap, $1024 \times 768$ pixels, anonymized filenames, no title) is identical to the one described in Sect.~\ref{sec:simulation}.

Six operations are applied. First, a 2 s window is extracted from the 6.04 s file. For positives the window is centered not on the catalogued time of arrival but on the midpoint of the dispersive sweep, computed from the catalogued DM, so that the whole track falls inside the frame; all 600 sweeps were verified to be fully contained, with a minimum margin of 0.35 s to the frame edge. For negatives, which have no time of arrival, the window is placed pseudo-randomly under a fixed seed. Second, the PSRFITS is decoded applying the \texttt{DAT\_SCL}, \texttt{DAT\_OFFS}, and \texttt{ZERO\_OFF} scalings, averaging the polarizations, and reordering the band in increasing frequency. Third, each channel is normalized by its median and a clipped standard deviation, which removes the bandpass. Fourth, dead and persistently corrupted channels are masked, using the \texttt{DAT\_WTS} weights, a robust $3\sigma$ criterion, and the exclusion of 410 channels at each band edge, which restricts the image to the useful 1050--1450 MHz of the receiver. Fifth, the block is averaged down to 1024 time bins by 256 frequency channels, which raises the signal-to-noise per pixel for bursts broader than the resulting 1.97 ms bin while diluting the narrowest ones; the catalogued widths of FAST-FREX span 0.34 to 78.52 ms~\cite{guo2025accelerating}, so both regimes are present, and this decimation is itself part of what limits the recall analyzed in Sect.~\ref{sec:reallimit}. Sixth, each row of the decimated image is renormalized, which flattens the residual variance correlated between neighboring channels, characteristic of the polyphase filterbank, and erases residual lines of persistent interference while preserving transients localized in time. Across the 1600 files, a median of 1218 of the 4096 channels was masked.

\begin{figure*}
\includegraphics[width=\textwidth]{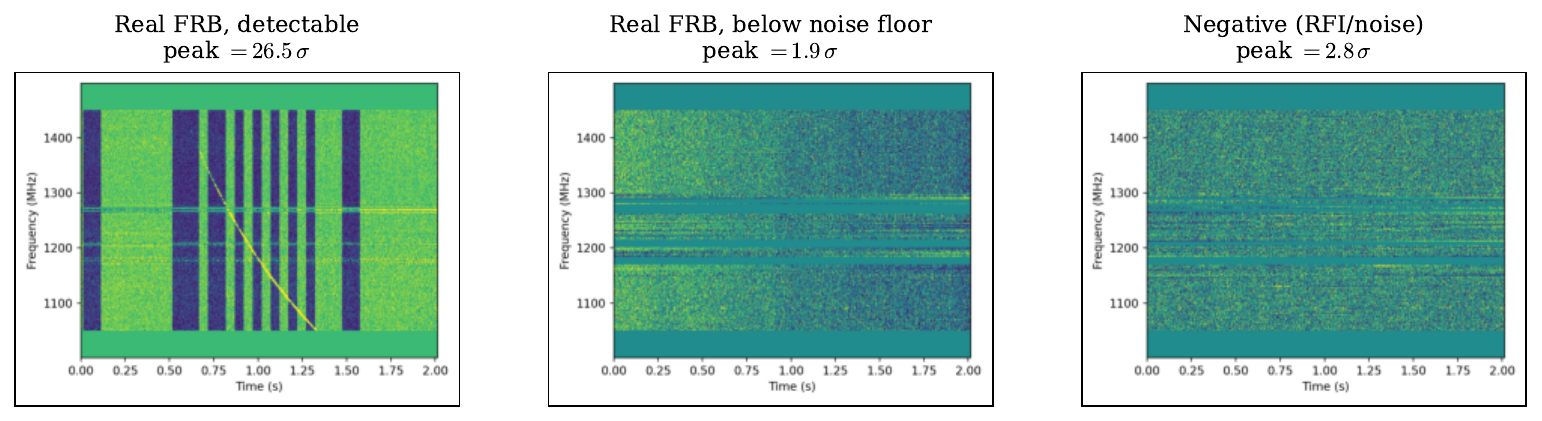}
\caption{\label{fig:examplesreal}Three real anonymized dynamic spectra from FAST-FREX, as supplied to the VLM. Left: a catalogued burst whose dispersive sweep is plainly visible, together with the broadband blanking and persistent lines typical of a real observation. Middle: a catalogued burst of the same data set whose dedispersed peak does not rise above the noise. Right: a negative sample. The middle and right panels are visually indistinguishable, although the middle one is a labeled FRB; the quantity reported above each panel is the dedispersed peak significance defined in Sect.~\ref{sec:reallimit}.}
\end{figure*}

The burst is deliberately \emph{not} dedispersed. The dispersive sweep is the visual signature that the prompt of Appendix~\ref{app:prompt} defines as an FRB, and an undispersed vertical pulse is explicitly to be classified as \texttt{NON\_FRB}; dedispersing the data would destroy the very feature on which the decision rests. Figure~\ref{fig:examplesreal} shows three resulting images, exactly as supplied to the model.

\subsection{Zero-shot performance on real bursts}\label{sec:realresults}

Both models classified all 1600 images, producing 3200 inferences with no parsing failure and no execution error, at a mean of 2.89 s per image for Gemma 4 E2B and 5.12 s for Gemma 4 E4B, with medians of 2.84 s and 5.09 s, slightly above the means of 2.7 s and 4.9 s reported in Sect.~\ref{sec:cost} for the simulated benchmark. Two caveats keep these measurements from being read as an operational throughput. They cover the model call alone, so the preprocessing of Sect.~\ref{sec:realprep}, which the simulated path does not require, is not included; and each FAST-FREX file contains 6.04 s of observation, of which only a 2 s window is classified. Even measured against that 2 s window alone, neither model runs faster than real time with the full binary prompt, as was also the case on the simulated benchmark. The discrete metrics are reported in Table~\ref{tab:realbinary} and the confusion matrices in Table~\ref{tab:realconfusion}.

\begin{table*}
\caption{\label{tab:realbinary}Discrete metrics on the full FAST-FREX data set (600 real FRBs, 1000 real negatives) at the default operating threshold of 0.5, with the same prompt and decoding configuration used in Sect.~\ref{sec:binary}. Intervals in brackets are 95\% Wilson intervals, defined in Sect.~\ref{sec:metrics}; the one on precision is conditional on the number of positive predictions the model made, 173 and 167, which is itself a random quantity rather than a design parameter.}
\begin{ruledtabular}
\begin{tabular}{lcc}
Metric & Gemma 4 E2B & Gemma 4 E4B \\
\hline
Recall (\texttt{FRB}) & 0.2850 [0.250, 0.322] & 0.2700 [0.236, 0.307] \\
Specificity & 0.9980 [0.993, 0.999] & 0.9950 [0.988, 0.998] \\
Precision (\texttt{FRB}) & 0.9884 [0.959, 0.997] & 0.9701 [0.932, 0.987] \\
F1-score (\texttt{FRB}) & 0.4424 & 0.4224 \\
F1-score (\texttt{NON\_FRB}) & 0.8224 & 0.8179 \\
Macro F1-score & 0.6324 & 0.6202 \\
Accuracy & 0.7306 & 0.7231 \\
\end{tabular}
\end{ruledtabular}
\end{table*}

\begin{table}
\caption{\label{tab:realconfusion}Confusion matrices on the full FAST-FREX data set at the 0.5 threshold, with \texttt{FRB} as the positive class. TP, FN, FP, and TN are defined in Sect.~\ref{sec:metrics}.}
\begin{ruledtabular}
\begin{tabular}{lcccc}
Model & TP & FN & FP & TN \\
\hline
Gemma 4 E2B & 171 & 429 & 2 & 998 \\
Gemma 4 E4B & 162 & 438 & 5 & 995 \\
\end{tabular}
\end{ruledtabular}
\end{table}

The two halves of the result move in opposite directions. On the negative class the models behave better than on the simulated benchmark, with 2 false positives in 1000 real samples of interference and noise for Gemma 4 E2B and 5 for Gemma 4 E4B, against 24 and 23 in the 1000 simulated \texttt{NON\_FRB} samples of Table~\ref{tab:fp}, that is, specificities of 0.9980 and 0.9950 against 0.9760 and 0.9770. Real interference, with all its instrumental structure, dead channels, and persistent lines, is rejected at least as reliably as the synthetic interference of the benchmark, by models never trained on this task. On the positive class, however, the recall falls from 0.9050 and 0.8240 on simulated FRBs to 0.2850 and 0.2700 on real ones, a drop of 62 and 55 percentage points.

The two models are also far more alike here than in simulation. They agree on 1562 of the 1600 images, and although Gemma 4 E2B is ahead of Gemma 4 E4B on every aggregate metric, the difference is not significant. Of the 38 discordant images, 25 favor the smaller model and 13 the larger, which a McNemar test does not separate from chance ($p = 0.07$). The direction nonetheless matches the simulated binary task, where the same interaction between a conservative scoring profile and the fixed threshold described in Sect.~\ref{sec:binary} favors the smaller model, and the multiclass ordering of Sect.~\ref{sec:multiclass-results}, in which Gemma 4 E4B is consistently the better of the two, is untouched by this experiment. The quantization of the score noted in Sect.~\ref{sec:prob} is, if anything, more extreme on real data. Gemma 4 E2B emitted only three distinct values of \texttt{frb\_probability} over the whole set, and Gemma 4 E4B nine, with 1427 of its 1600 responses concentrated on a single value.

\subsection{What limits the recall}\label{sec:reallimit}

Taken alone, a recall of 0.28 would suggest that the models fail on real bursts. That reading is incomplete, and the reason is that a catalogued burst is not necessarily a burst that is visible in a 2 s full-band image. To separate the two, each image was dedispersed at the catalogued DM of its sample, without wrap-around, and the resulting time profile was searched with a boxcar matched filter over widths from 1 to 32 bins; the peak of that profile, in units of a noise scale estimated from the median absolute deviation of the profile itself, measures how much dispersed signal the image actually contains. For the negatives, which have no catalogued DM, the same quantity was computed in two ways, because the number of trials is what sets the height of a noise floor. A single trial DM, drawn from the distribution of the catalogued ones, gives each negative exactly the number of trials a positive receives, one DM and six boxcar widths, and therefore a floor that can be compared with the positives directly. A blind grid of 24 trial DMs from 100 to 700 pc cm$^{-3}$, maximized over the grid, instead answers what an untargeted search would face; since the maximum of 144 noise trials exceeds the maximum of 6, this second floor is higher by construction, and both are reported. As a validation, the recovered peak time was compared with the catalogued time of arrival for all 600 positives, and the median discrepancy is 2.1 ms, about one decimated time bin, 90\% of the bursts agree to within 10 ms, and 96\% within 20 ms. The diagnostic is therefore recovering the catalogued burst and not an unrelated feature of the image.

\begin{figure}
\includegraphics[width=0.49\textwidth]{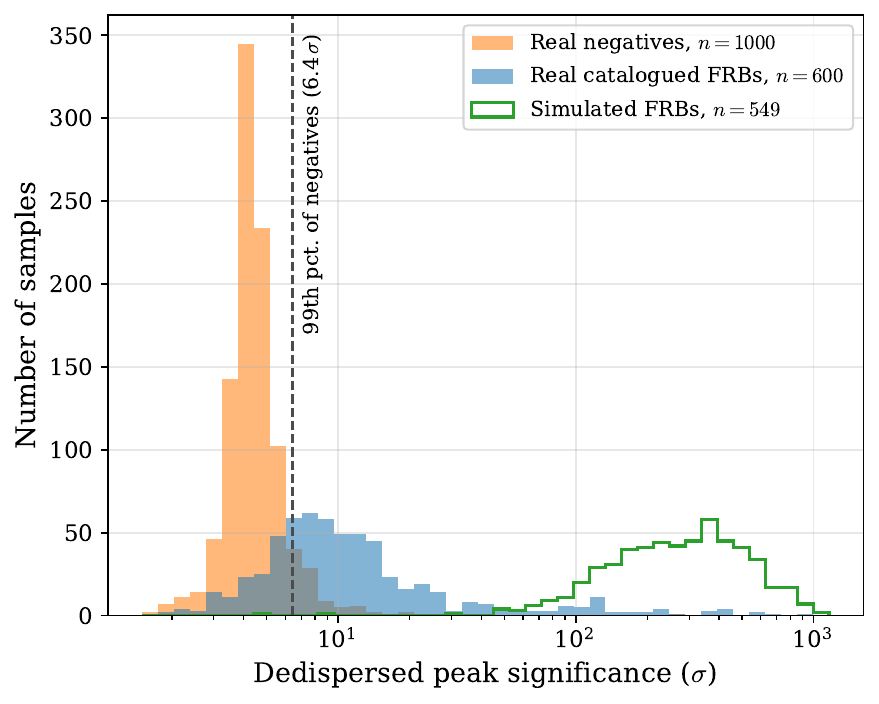}
\caption{\label{fig:snrhist}Distribution of the dedispersed peak significance for the 600 catalogued FRBs and the 1000 negatives of FAST-FREX, on a logarithmic scale, together with the 549 simulated bursts of Sect.~\ref{sec:simulation} whose sweep is fully contained in the file, measured on the same time--frequency grid. The dashed line marks the 99th percentile of the negatives measured with the same number of trials as a positive; 25\% of the catalogued bursts fall below it. The simulated population is separated from the real one by more than an order of magnitude in significance, which is what makes the recall gap between the two benchmarks interpretable.}
\end{figure}

Figure~\ref{fig:snrhist} shows the resulting distributions. They overlap heavily. The median peak is $9.4\sigma$ for the 600 catalogued bursts and $3.3\sigma$ for the 1000 negatives measured with a single trial; the 99th percentile of that floor is $6.4\sigma$, and 152 of the 600 positives, that is 25\%, fall below it. Measured against the blind floor, whose 99th percentile rises to $11.5\sigma$ precisely because it is maximized over 24 trial DMs, the fraction is 61\%. The first figure is the fair comparison between the two classes and the one to quote; the second is what an untargeted search would confront, and it is an upper bound. On either measure a substantial part of the catalogued positives carries no more dispersed signal, in the representation the model receives, than interference and noise produce on their own.

\begin{figure}
\includegraphics[width=0.49\textwidth]{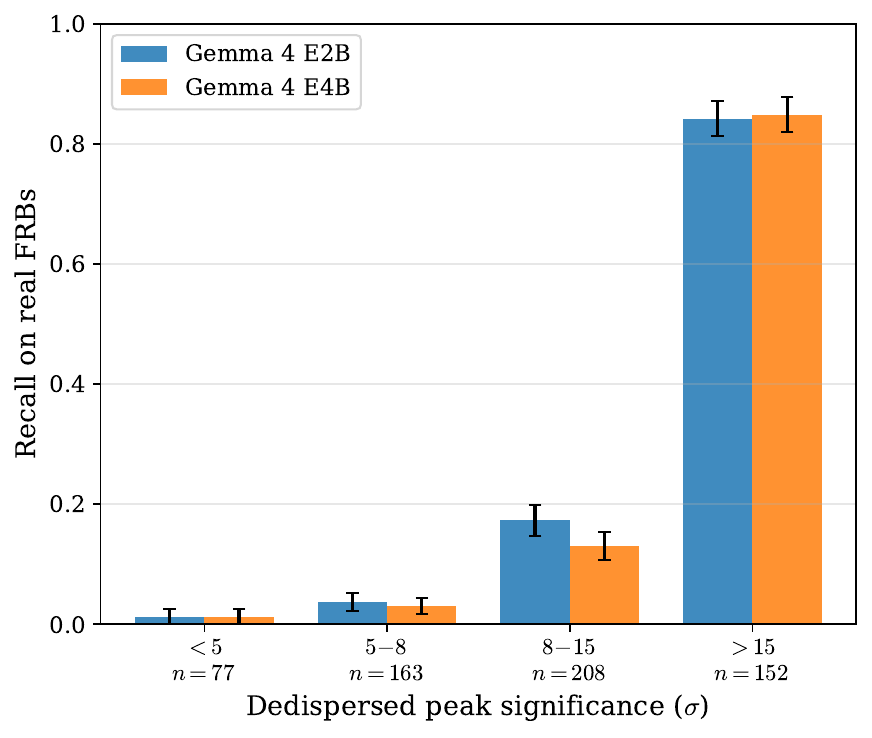}
\caption{\label{fig:recallsnr}Recall of both VLMs on real FRBs as a function of the dedispersed peak significance of the burst. Error bars are binomial standard errors and $n$ is the number of catalogued bursts in each band. Below $5\sigma$ each model recovers 1 of the 77 bursts; above $15\sigma$ both recover 84 and 85\% of them.}
\end{figure}

Once the positives are stratified by this quantity, the behavior of the models becomes monotonic and easy to read, as shown in Fig.~\ref{fig:recallsnr}. Below $5\sigma$, where 77 bursts fall, each model recovers a single one, a recall of 0.013. Between 5 and $8\sigma$ the recall is 0.037 and 0.031, between 8 and $15\sigma$ it is 0.173 and 0.130, and above $15\sigma$, where 152 bursts fall, it rises to 0.842 and 0.849. The stratification depends only on the image and not on the classifier, so the same 152 bursts are scored for both models; of them, 123 are recovered by both, 5 by Gemma 4 E2B alone, 6 by Gemma 4 E4B alone, and 18 by neither.

This is the central result of the section, and the simulated benchmark makes it quantitative rather than suggestive. Applying the same diagnostic to the injected bursts of Sect.~\ref{sec:simulation}, decimated beforehand to the same 1.97 ms by 256-channel grid so that the two significances measure the same physical quantity, gives a median peak of $277\sigma$ against $9.4\sigma$ for the real catalogued bursts, a factor of 29.5. Of the simulated bursts whose sweep is fully contained in the 2 s file, 547 of 549 lie above $15\sigma$ and exactly one falls below the 99th percentile of pure simulated noise, against 152 of 600 above $15\sigma$ in FAST-FREX. That the benchmark injects its bursts well above the noise is therefore a measurement and not an assumption, and the two data sets are seen to occupy almost disjoint regimes of detectability.

The comparison of recall can then be made at matched brightness. Restricted to those same 549 simulated samples the models reach 0.927 and 0.874, against 0.842 and 0.849 on the real bursts above $15\sigma$. The gap that the raw figures suggest, 0.905 and 0.824 on simulated data against 0.285 and 0.270 on real data, therefore closes almost entirely once the difference in burst detectability is removed, and what remains is small enough to be plausibly accounted for by the residual differences between the two paths, since the real images carry RFI excision, masked channels, and a real bandpass, none of which the simulated ones require. The global recall on real data is thus governed by the composition of FAST-FREX rather than by a failure of visual recognition, since only 152 of its 600 positives are unambiguous in a 2 s undedispersed full-band image, and the models find most of those.

The same effect explains the differences between the three sources, reported in Table~\ref{tab:realsource}. The recall on FRB20201124 is more than three times that on FRB20121102, and the confidence intervals do not overlap, but the bursts of FRB20201124 are also brighter, with a median peak of $22.2\sigma$ against $8.7\sigma$. Controlling for brightness reduces the gap considerably without eliminating it. Among bursts above $15\sigma$, Gemma 4 E2B recovers 54 of 71 from FRB20121102 and 71 of 78 from FRB20201124. A residual difference therefore remains once brightness is accounted for, plausibly related to spectral occupancy or burst width, which the present diagnostic does not measure. Since FRB20121102 alone contributes 470 of the 600 positives, the global recall is in practice dominated by the faintest of the three sources; a data set with a different source composition would yield a different figure for the same models.

\begin{table*}
\caption{\label{tab:realsource}Recall by source on FAST-FREX, with 95\% Wilson intervals, together with the median catalogued DM (in pc cm$^{-3}$) and the median dedispersed peak significance of each source. FRB20180301 contributes only 5 bursts and supports no conclusion.}
\begin{ruledtabular}
\begin{tabular}{lccccc}
Source & $n$ & Median DM & Median peak & Recall, Gemma 4 E2B & Recall, Gemma 4 E4B \\
\hline
FRB20121102 & 470 & 566.0 & $8.7\sigma$ & 0.187 [0.15, 0.22] & 0.166 [0.14, 0.20] \\
FRB20180301 & 5 & 517.8 & $19.0\sigma$ & 0.800 [0.38, 0.96] & 0.800 [0.38, 0.96] \\
FRB20201124 & 125 & 412.6 & $22.2\sigma$ & 0.632 [0.54, 0.71] & 0.640 [0.55, 0.72] \\
\end{tabular}
\end{ruledtabular}
\end{table*}

Four caveats bound this analysis. The dedispersed peak is a diagnostic defined here, not a quantity published with the data set, and it integrates the full band without matching the spectral extent of the burst, so it underestimates the significance of band-limited emission. The window for the positives is centered using the catalogued DM and time of arrival, which makes this a candidate-classification experiment and not a blind search; the recall reported here is an upper bound on what the same models would achieve when scanning an unsegmented observation. The comparison with the simulated benchmark is restricted to the 549 injected bursts whose sweep is fully contained in the file, because with DMs drawn up to $899$ pc cm$^{-3}$ the sweep across the L band reaches 2.07 s, longer than the file itself, and a truncated dedispersion cannot integrate what has left the frame. Those 451 remaining bursts are not invisible --- the models recover 88\% and 76\% of them, since a sweep crossing the whole frame is conspicuous --- but the diagnostic cannot assign them a significance, and they are excluded rather than patched. Finally, the three sources are repeaters observed with a single telescope, so the result characterizes this data set rather than the FRB population. This last restriction compounds the first. Machine-learning analyses of the CHIME/FRB catalogs consistently find that repeaters are biased toward narrowband emission while apparently non-repeating bursts are broader in band~\cite{sun2025exploring,sun2026practical,sun2026unveiling}, so a sample composed entirely of repeaters is enriched in precisely the band-limited emission that a full-band view represents least favorably. The point is not only statistical. The two sources that supply 595 of the 600 FAST-FREX positives, FRB20121102 and FRB20201124, are both placed by that analysis in its highest-repetition and narrowest-band region of the spectral-index--spectral-running plane~\cite{sun2026practical}, so the bias applies to this data set specifically and not merely to repeaters as a class. A sample of one-off bursts would be expected to populate the detectable end of Fig.~\ref{fig:snrhist} more densely at the same intrinsic brightness. This last step carries an assumption of its own, since those analyses rest on CHIME observations at 400--800 MHz while FAST-FREX is an L-band data set, and the spectral extent of a burst need not be preserved across instruments and observing bands.

What the section establishes, then, is narrower and firmer than a single accuracy figure. Under a prompt-only regime, with no exposure to real data of any kind, both models reject real interference almost perfectly and recognize the majority of the real bursts that their input representation actually renders visible. The bottleneck is the representation, not the classifier, which points the next step at the input rather than at the model, namely sub-band views, time--DM diagnostics, and dedispersed candidate cutouts, of the kind that the specialized pipelines of Sect.~\ref{sec:frb} construct before classification.

\subsection{Comparison with specialized detectors on the same data set}\label{sec:realswin}

FAST-FREX is a shared benchmark, and the detectors of Sect.~ \ref{sec:frb} have published results on the very files evaluated here. That makes it possible to place the zero-shot generalist among them without re-running anything, and Table~\ref{tab:realswin} does so. The comparison is with published values rather than a paired experiment of the kind performed against SwinYNet on simulated data in Sect.~ \ref{sec:paired}, and the two sides do not read the files in the same way: the specialized detectors search the full 6.04 s recording at native resolution, whereas the VLMs receive the 2 s decimated window of Sect.~\ref{sec:realprep}, centered using the catalogued DM, an asymmetry that favors the VLMs, so the recall gap is not an artifact of it.

\begin{table*}
\caption{\label{tab:realswin}Detection performance on the 600 catalogued bursts and 1000 negatives of FAST-FREX. The first five rows are the published values of Table 2 of Ref.~\cite{chen2026swinynet}, obtained by their authors with their own preprocessing; the last two are the zero-shot VLMs of this work. The false positives of PRESTO and Heimdall are candidate-level rather than file-level, which is why their precision falls orders of magnitude below the rest; for SwinYNet, RaSPDAM, DRAFTS, and the two VLMs the count is the number of negative files declared positive, and is directly comparable.}
\begin{ruledtabular}
\begin{tabular}{lccccc}
Method & Input & TP & FP & Recall & Precision \\
\hline
PRESTO ($\mathrm{S/N}=5$) & FITS & 513 & 17406 & 0.855 & 0.028 \\
Heimdall & FITS & 489 & 5854 & 0.815 & 0.077 \\
RaSPDAM & FITS & 501 & 14 & 0.835 & 0.973 \\
DRAFTS & FITS & 580 & 23 & 0.967 & 0.962 \\
SwinYNet & FITS & 574 & 0 & 0.957 & 1.000 \\
Gemma 4 E2B, zero-shot & PNG & 171 & 2 & 0.285 & 0.988 \\
Gemma 4 E4B, zero-shot & PNG & 162 & 5 & 0.270 & 0.970 \\
\end{tabular}
\end{ruledtabular}
\end{table*}

The result splits along the two axes rather than along a single ranking. On precision the zero-shot models are competitive with the trained detectors, 0.988 and 0.970 against 0.973 for RaSPDAM and 0.962 for DRAFTS, and only SwinYNet, which produced no file-level false positive at all, is clearly ahead. On recall they are not competitive at all, 0.285 and 0.270 against 0.835 to 0.967. Generalist pretraining thus supplies enough to rule out a real interference pattern, but not enough to recover most bursts from the representation the model is given.

That second half deserves to be stated plainly, because the simulated benchmark of Sect.~\ref{sec:results} does not anticipate it. There, Gemma 4 E2B was statistically indistinguishable from SwinYNet at the default threshold. On real data the same two systems are separated by a factor of more than three in recall. The proximity measured on simulated data does not transfer, and any operational reading should rest on Table~\ref{tab:realswin} rather than on Table~\ref{tab:binary}.

The stratified results of Ref.~\cite{chen2026swinynet} sharpen the diagnosis of Sect.~\ref{sec:reallimit} rather than contradicting it. Their Table 3 reports a recall of 0.927 in their faintest signal-to-noise interval and 1.000 in their brightest, against the 0.013 we obtain below $5\sigma$. Their significance is not our $\sigma$, being computed from the catalogued burst parameters rather than from the delivered image, so the bands are not numerically comparable. The qualitative statement is nonetheless unambiguous. A trained detector reading the full 6.04 s file at native resolution recovers the great majority of exactly those bursts that our 2 s decimated full-band image leaves at the noise floor. The signal is present in the data and is recoverable; what discards it is the representation, and this is now established by an independent system rather than by our own diagnostic. It is worth noting that their recall is not monotonic in signal-to-noise, a behavior their authors call counterintuitive and attribute to local saliency, whereas ours is strictly monotonic in $\sigma$. The two systems therefore fail for different reasons, which is itself an argument for treating them as complementary rather than as competitors.

\subsection{Threats to validity}\label{sec:threats}

Some limitations must be made explicit so that the results are interpreted correctly.

First, the main benchmark is simulated. This ensures label control, balancing, and reproducibility, but it does not capture the full complexity of real observational data, including instrumental effects, RFI variability, calibration, masking, defective channels, and real signal-to-noise distributions. This is the limitation that Sect.~\ref{sec:real} addressed, evaluating the same models, unchanged, on the 1600 real FAST observations of FAST-FREX~\cite{guo2025accelerating}, and it is on those results that any operational reading should rest. That validation carries its own restrictions, set out in Sect.~\ref{sec:reallimit}. The window of each positive is centered using the catalogued DM and time of arrival, which makes it a candidate-classification experiment rather than a blind search, and its three sources are repeaters observed with a single telescope.

Second, the comparison between the VLM and SwinYNet is asymmetric. SwinYNet was developed as a specialized detector and operates on FITS, while the VLM receives only a PNG image. This asymmetry favors the specialized detector in terms of access to information and should be seen as part of the experimental design. Precisely for this reason, the performance of Gemma 4 E2B, so close to SwinYNet at the default threshold, is notable, but it should not be described as proof of general superiority.

Third, the simulated data set is balanced. This choice is appropriate for an initial controlled comparison, but it does not represent the real prevalence of FRBs in astronomical searches. In real scenarios, even a small false-positive rate can generate many false candidates. For this reason, Precision-Recall curves, threshold analyses, and evaluations under rare priors must be discussed before operational use is proposed. The real-data evaluation of Sect.~\ref{sec:real} is not balanced, its 1000 negatives against 600 positives placing an empirical bound of a few per mille on the false-positive rate, but it is still far from the prevalence of a production stream.

Fourth, the VLM depends on the visual representation. Colormap, normalization, resolution, contrast, and cropping can affect the decision. The result presented here holds for the image protocol described in Sect.~\ref{sec:simulation}. Future studies should test visualization ablations~\cite{meyes2019ablation, sheikholeslami2019ablation} and compare the robustness to different ways of representing the same FITS. The same care applies to the design of the textual prompt, since the outputs of generalist VLMs can vary with changes in the structure of the instruction~\cite{drozdova2025radio}. As discussed in Sect.~\ref{sec:decoding}, deterministic decoding removes only the stochastic variation, not this dependence on the prompt, so that the results hold for the fixed prompt reproduced in Appendix~\ref{app:prompt}.

Fifth, the probabilities produced by the VLMs in this run are quantized into few distinct values (three for Gemma 4 E2B, eight for Gemma 4 E4B), and the same holds on real data, where the two models emitted three and nine distinct values over 1600 images. This makes the probabilistic and calibration metrics coarse and limits the ROC and Precision-Recall curves to few operating points. As indicated in Sect.~\ref{sec:prob}, obtaining a finer continuous score is the next methodological step to make this part of the evaluation conclusive.

Sixth, the evaluation is zero-shot with respect to this benchmark, in the sense that no labeled example from it is shown to the models, but the data used to pretrain generalist VLMs may include dynamic-spectrum figures of FRBs from the scientific literature. Some indirect prior exposure to the visual concept of a dispersed burst therefore cannot be excluded. This does not affect the paired comparison on these newly simulated images, but it should temper any reading of the results as evidence that the models had no previous contact with the domain.

Finally, the textual justifications of the VLM are useful for auditing, but they should not be automatically treated as correct physical explanations. A model can produce a plausible justification for a wrong decision. Thus, interpretability here should be understood as inspectability of the decision, not as a guarantee of physical causality.

\section{Conclusions}\label{sec:conclusions}

This work shows that generalist, small, locally run VLMs, operated with a prompt alone and no task-specific training, already perform scientifically useful FRB triage, and that their failure modes are structured, complementary to those of a dedicated detector, and traceable to the input representation rather than to the classifier.

In the \texttt{FRB} versus \texttt{NON\_FRB} task, Gemma 4 E2B and E4B achieved an accuracy of 0.90--0.94 and a ROC-AUC of 0.92--0.95, with Gemma 4 E2B showing no statistically significant difference from a specialized detector at the default threshold and rejecting RFI significantly more efficiently. Moreover, the same model proved capable of going beyond the binary decision and separating \texttt{FRB}, \texttt{RFI}, and \texttt{NOISE} only by changing the prompt.

SwinYNet remains the performance reference, with a perfect probabilistic ranking on this benchmark. It should be read as the ceiling of a dedicated detector, one that a zero-shot generalist model approaches to within a few points of ROC-AUC on simulated data. That proximity does not survive the move to real observations. On the same 600 catalogued bursts of FAST-FREX, the published results of Ref.~\cite{chen2026swinynet} give SwinYNet a recall of 0.957 with no file-level false positive, against 0.285 and 0.270 here, so the two systems are separated on real data by a factor of more than three in recall while remaining comparable in precision. Section~\ref{sec:realswin} sets out that comparison and its limits, and it is the one an operational reading should use.

The two halves of that real-data result separate cleanly into what generalist pretraining does and does not provide. On the 1000 real negatives of FAST-FREX the models produced 2 and 5 false positives, a rejection of genuine instrumental interference at least as reliable as on the synthetic RFI of the simulated benchmark and obtained without ever having seen real data. Stratifying the positives by the dispersed signal their images actually contain resolves the apparent contradiction. A quarter of the catalogued positives carry less signal than one in a hundred negatives produces by chance, each model recovers 1 of the 77 bursts that fall below $5\sigma$, and both recover 84 and 85\% of the 152 whose sweep is unambiguous. Applying the same diagnostic to the simulated benchmark, on the same time--frequency grid, shows why the two figures differed in the first place. Its injected bursts have a median significance nearly 30 times higher, and essentially all of them sit in the band where the real recall already stands at 84\%. Matched for brightness, the simulated recalls of 0.927 and 0.874 stand against real recalls of 0.842 and 0.849. The limiting factor on real data is therefore the visual representation supplied to the model, not its capacity to recognize a dispersed sweep, and the natural next step lies at the input, in sub-band views, time--DM diagnostics, and dedispersed candidate cutouts, rather than in a larger model.

The value of this work is not in declaring a winner, but in demonstrating that a generalist, inexpensive, and interpretable tool already recognizes FRB, RFI, and noise patterns with little domain information. In the multiclass configuration this recognition already runs faster than real time at the inference stage, with each 2 s spectrum classified in 1.0--1.5 s on a single workstation; in a controlled 200-image subsample, a leaner binary prompt that adds a one-sentence justification before the decision matches or exceeds the accuracy of the full binary prompt at 1.16 s (Gemma 4 E2B) and 1.78 s (Gemma 4 E4B) per spectrum, bringing the binary task itself below the observation time for both models. This opens the way for applications in radio astronomy (triage, diagnosis, and the detection of signal types not anticipated in the training of specialized models), and motivates the next steps, such as a finer probabilistic score, DM estimation, richer input representations for real observations, and a blind search over unsegmented data, the real-data evaluation of Sect.~\ref{sec:real} being a candidate-classification experiment rather than a search. In addition, the adaptation techniques deliberately left out of this study --- in-context (few-shot) learning, fine-tuning, retrieval-augmented generation (RAG), and agents or calls to external tools --- can still be employed to further improve the results. A further direction is suggested by the morphological division of the population itself. Since repeating and apparently non-repeating bursts are separable largely by spectral morphology~\cite{sun2025exploring,sun2026practical,sun2026unveiling}, and morphology is what the model reads from the image, the same prompt-only mechanism that reconfigures the classifier from binary to three-class could be pointed at that distinction.

\begin{acknowledgments}
The work of A.R.Q. is supported by FAPESQ-PB. A.R.Q. acknowledges support by CNPq under process number 306884/2026-7. R.A.B., T.S.S.D. thank the Paraíba State Research Support Foundation (FAPESQ) for financial support. The work of R.H.S. and K.E.L.F. is supported by CNPq. K.E.L.F. also thanks FAPESQ for the exchange program Paraíba sem Fronteiras and the ENSEMBLE3 project, which is carried out within the 2.1 International Research Agendas programme of the Foundation for Polish Science co-financed by the European Union under the European Funds for Smart Economy 2021-2027 (FENG.02.01-IP.05-0044/24), project (MAB/2020/14), which is carried out within the International Research Agendas Programme (IRAP) of the Foundation for Polish Science co-financed by the European Union under the European Regional Development Fund and the Teaming Horizon 2020 programme (GA. No. 857543) of the European Commission and the project of the Minister of Science and Higher Education "Support for the activities of Centers of Excellence established in Poland under the Horizon 2020 program" under contract No. MEiN/2023/DIR/3797. 
\end{acknowledgments}

\section*{Data availability}

The code that generates, classifies, and evaluates the samples, together with the benchmark metadata (the label and image manifests), the per-sample predictions of the three systems, and the evaluation outputs, is openly available in the project repository at \url{https://github.com/raiffhugo/frb-vlm-benchmark}. The simulated PSRFITS files and the rendered images are too large to be hosted in the repository (about 60~GB); they can be regenerated exactly from the pinned configuration and global seed described in Sect.~\ref{sec:simulation}, and are also available from the corresponding author upon reasonable request.

The real observations analyzed in Sect.~\ref{sec:real} are the public FAST-FREX data set~\cite{guo2025accelerating}, distributed by the Science Data Bank under DOI 10.57760/sciencedb.15070 (about 442~GB). The repository includes the code that selects and retrieves those files, the preprocessing stage that converts them into the images supplied to the model, the resulting image manifest with the window and masking parameters recorded per sample, the per-sample predictions of both models, and the dedispersed-peak diagnostic used in Sect.~\ref{sec:reallimit}.

The pipeline is implemented in Python and builds on open-source software: simulateSearch~\cite{Luo2022_simulateSearch_paper, hobbs2022simulatesearch}, Hugging Face Transformers~\cite{wolf2020transformers}, PyTorch~\cite{paszke2019pytorch}, NumPy~\cite{harris2020array}, scikit-learn~\cite{pedregosa2011scikit}, Matplotlib~\cite{hunter2007matplotlib}, and Astropy~\cite{astropy2022}.

\appendix

\section{Binary classification prompt}\label{app:prompt}

The full binary prompt supplied to the model in the \texttt{FRB} versus \texttt{NON\_FRB} task is reproduced below, exactly as sent to the model.

\onecolumngrid
\begin{lstlisting}
You are an expert radio astronomer classifying dynamic-spectrum images from a transient-search pipeline.

Image meaning:
- x-axis: time
- y-axis: frequency
- intensity: signal power

Classify the image as exactly one of:
- FRB
- NON_FRB

Return ONLY valid JSON. No markdown, no comments, no extra keys.

Decision rules:

1. FRB candidates
Label as FRB only when the dominant signal is a SINGLE localized transient that is plausibly broadband and shows a frequency-dependent arrival-time trend, usually later arrival at lower frequency. The sweep may be curved, nearly linear, faint, or partially masked.

2. Low-S/N sensitivity
Do not treat faint coherent structure as pure noise. If there is any plausible faint diagonal/curved sweep, localized broadband brightening, partial dispersed track, or excess variance along a tilted band, use an intermediate probability instead of a very low one.

3. NON_FRB cases
Label as NON_FRB when the image is dominated by:
- pure random background with no coherent transient,
- persistent horizontal frequency bands,
- persistent vertical time bands,
- periodic stripes or grid-like structure,
- zero-DM broadband vertical impulse with no dispersion delay,
- saturation, clipping, blank regions, uniform panels, or normalization artifacts,
- two or more temporally separated broadband pulses or repeating patterns.

4. Probability calibration
frb_probability must be in [0,1].
Use the full range and avoid always reusing the same values.

Calibration guide:
- 0.95-0.99: textbook FRB, clear broadband dispersed sweep.
- 0.80-0.94: probable FRB with minor ambiguity.
- 0.60-0.79: plausible FRB but faint, partial, or noisy.
- 0.45-0.59: genuinely ambiguous.
- 0.30-0.44: weak possible structure, more likely NON_FRB.
- 0.18-0.29: likely noise or weak contamination.
- 0.05-0.17: clear RFI, persistent bands, artifacts, or strong NON_FRB evidence.
- 0.01-0.04: blank, saturated, clipped, or unambiguous artifact.

Label consistency:
- label = "FRB" only if frb_probability >= 0.50
- label = "NON_FRB" only if frb_probability < 0.50

confidence is how certain you are about the discrete label, not the FRB probability.

Output exactly this JSON schema:

{
  "label": "FRB|NON_FRB",
  "frb_probability": 0.0,
  "confidence": 0.0,
  "reason": "short explanation referencing the visual evidence",
  "features": {
    "visible_structure": true,
    "broadband": true,
    "localized_transient": true,
    "frequency_dependent_delay": true,
    "persistent_bands": false,
    "repeating_pattern": false,
    "uniform_or_constant_background": false,
    "saturated_or_clipped": false,
    "random_background_texture": false
  }
}

Feature definitions:
- visible_structure: any coherent visual structure is present.
- broadband: signal spans a substantial frequency range.
- localized_transient: signal is confined to a short time interval.
- frequency_dependent_delay: arrival time changes systematically with frequency.
- persistent_bands: horizontal or vertical bands persist across much of the panel.
- repeating_pattern: multiple pulses, stripes, grids, or periodic structures.
- uniform_or_constant_background: panel is nearly blank or constant.
- saturated_or_clipped: image has saturation, clipping, or rectangular masked regions.
- random_background_texture: dominated by stochastic noise with no coherent candidate.

The reason must be one short sentence.
\end{lstlisting}
\twocolumngrid

\section{Multiclass classification prompt}\label{app:multiclass-prompt}

The prompt supplied to the model in the multiclass \texttt{FRB}/\texttt{RFI}/\texttt{NOISE} task is reproduced below, exactly as sent. In contrast to the binary prompt, it defines \texttt{RFI} as a residual class rather than enumerating interference morphologies, and requests a lighter JSON output without the probability field or the per-feature block.

\onecolumngrid
\begin{lstlisting}
You are an expert in radio astronomy analyzing candidate images from a transient-search pipeline.

The image is a dynamic spectrum: the horizontal axis is time, the vertical axis is frequency, and pixel intensity is signal power.

Classify the image into exactly one of three classes: FRB, RFI, or NOISE.
Decide only from the visual content of the image.

Class definitions
=================
NOISE
The entire panel is featureless, stochastic background texture. There is no coherent structure anywhere: no lines, no stripes, no bands, no localized bright spots, no sweep, nothing but random fluctuations.

FRB
A single astrophysical burst: one localized, broadband transient whose arrival time changes systematically (monotonically) with frequency, tracing a single continuous sweep across the band. It appears only once, is confined in time, and neither repeats nor sits at a fixed frequency.

RFI
Anything else. Any image that is not pure NOISE and is not a single dispersed FRB sweep is RFI, i.e. any coherent, persistent, repeated, or artificial-looking structure of terrestrial or instrumental origin.

Decision procedure
=================
1. Is there a single broadband sweep that drifts monotonically in frequency with time? If yes, label FRB.
2. Otherwise, is the panel pure random texture with no coherent feature at all? If yes, label NOISE.
3. Otherwise, label RFI.

Rules
=====
- NOISE requires a complete absence of coherent structure. If you can see ANY non-random feature anywhere in the panel -- even faint, even sparse, even a single thin line, a streak, or a few repeated marks -- the image is NOT NOISE.
- Faint or sparse structure still counts as structure.
- An FRB drifts in frequency and appears once. Signals that stay at a fixed frequency, form horizontal or vertical bands, or repeat in time are RFI, not FRB.
- When unsure between RFI and NOISE, choose RFI. When a single drifting broadband sweep is unsure between FRB and RFI, prefer FRB.

Output
======
Return ONLY a valid JSON object with this exact schema, and nothing else:
{
  "label": "FRB|RFI|NOISE",
  "confidence": 0.0,
  "reason": "one short sentence citing the visual evidence"
}
confidence is how certain you are about the chosen label.
Do not include markdown, code fences, comments, or any text outside the JSON object.
\end{lstlisting}
\twocolumngrid



\end{document}